\documentclass{article}

\PassOptionsToPackage{numbers, compress}{natbib}
\usepackage[main, final]{neurips_2026}

\usepackage[utf8]{inputenc} 
\usepackage[T1]{fontenc}    
\usepackage{hyperref}       
\usepackage{url}            
\usepackage{booktabs}       
\usepackage{amsfonts}       
\usepackage{nicefrac}       
\usepackage{microtype}      
\usepackage{xcolor}         
\usepackage{subcaption}
\usepackage{graphicx}
\usepackage{amsmath}
\usepackage{colortbl}
\usepackage{wrapfig}
\usepackage{cleveref}

\title{Neural‑Visual Decoding via Cognitive‑guided Adaptive Blurring and Information‑Constrained Alignment}

\author{%
  \begin{minipage}[t]{\textwidth}
    \centering
    \normalfont
    \textbf{Fan Yin}\textsuperscript{1}\thanks{Equal contribution.}\quad
    \textbf{Chuhang Zheng}\textsuperscript{2}\footnotemark[1]\quad
    \textbf{Peiliang Gong}\textsuperscript{3}\quad
    \textbf{Donghai Guan}\textsuperscript{1}\thanks{Corresponding author.}\quad
    \textbf{Qi Zhu}\textsuperscript{3}\footnotemark[2]\\[6pt]
    {\small
      \textsuperscript{1}College of Computer Science and Technology,\\
      Nanjing University of Aeronautics and Astronautics, Nanjing, China\\[3pt]
      \textsuperscript{2}The International Joint Institute of Tianjin University, Fuzhou,\\
      Tianjin University, Tianjin, China\\[3pt]
      \textsuperscript{3}College of Artificial Intelligence,\\
      Nanjing University of Aeronautics and Astronautics, Nanjing, China\\[5pt]
      \texttt{fanyin@nuaa.edu.cn}\quad
      \texttt{chzheng@tju.edu.cn}\\
      \texttt{plgong@nuaa.edu.cn}\quad
      \texttt{dhguan@nuaa.edu.cn}\quad
      \texttt{zhuqi@nuaa.edu.cn}
    }
  \end{minipage}
}

\begin{document}

\maketitle

\begin{abstract}
  EEG-based visual decoding aims to establish a mapping between neural signals and visual semantics. However, it remains constrained by the dual challenges of severe information granularity mismatch and the low signal-to-noise ratio (SNR) of EEG signals. Existing approaches typically treat visual features as static, ignoring the dynamic selectivity of human vision and the frequency specificity of neural oscillations. To bridge this gap, we propose \textbf{CAIA}, a Cognitive-guided Adaptive blurring with Information-Constrained Alignment framework for Neural-Visual decoding. On the visual side, it simulates selective attention to adaptively reduce redundancy. Meanwhile, on the EEG side, it combines neural oscillation priors with an information-bottleneck-inspired spectral sparsity regularizer to enhance SNR. Specifically, we devise a cognitive-dynamics-based adaptive blurring mechanism that dynamically integrates center-biased and saliency-guided visual cues via cross-modal attention. Furthermore, we introduce a distribution-aware boundary calibration loss to robustly rectify alignment bias caused by outlier samples. Moreover, a cognitively guided information-screening method is proposed to select task-relevant EEG oscillations. Extensive experiments demonstrate that CAIA improves both subject-dependent and subject-independent average Top-1 and Top-5 accuracy in zero-shot brain-to-image retrieval, significantly outperforming prior methods. Our work validates that optimizing visual information density to match neural granularity offers a more interpretable and robust pathway for neural decoding. Code is available at https://github.com/mikumafu/CAIA.
\end{abstract}

\section{Introduction}
Understanding the brain's visual processing mechanism and decoding its neural signals is a cutting-edge research topic bridging neuroscience, cognitive science, and artificial intelligence \cite{t1,t2}. Establishing accurate stimulus-activity mappings advances brain-inspired intelligence and supports applications like Brain Computer Interfaces (BCIs) and visual prosthetics \cite{t3,t4,t5}. While deep learning has improved neural decoding, EEG's low SNR and cross-modal granularity mismatch hinder zero-shot generalization \cite{t6,t7}.

Functional magnetic resonance imaging (fMRI), magnetoencephalography (MEG), and electroencephalography (EEG) constitute the dominant neuroimaging modalities for visual neural decoding \cite{t8,t9}. While fMRI excels in spatial resolution, its temporal precision is fundamentally constrained by hemodynamic latency \cite{t10}. MEG affords millisecond-scale temporal fidelity but suffers from prohibitive costs and noise-sensitivity \cite{t11}. EEG is portable and temporally precise, ideal for real-time applications \cite{t12,t13}, yet suffers from low SNR due to noise and individual variability \cite{t14}. Adaptively extracting visual-related neural features remains a core challenge.

To bridge the representational gap, some studies focus on designing more efficient EEG encoders. For instance, BrainVis \cite{t15} hierarchically fuses spatiotemporal features via temporal self-supervised encoding and frequency-domain LSTM. NICE \cite{t16} employs a spatial attention module to model inter-electrode correlations. ATM \cite{t17} combines spatiotemporal convolution with channel-wise attention to extract multi-domain EEG features. These methods essentially attempt to distill discriminative patterns from noisy data through complex encoders. 

\begin{wrapfigure}{r}{0.55\textwidth}
  \centering
  \vspace{-8pt}
  \begin{subfigure}[t]{0.49\linewidth}
    \centering
    \includegraphics[width=\linewidth]{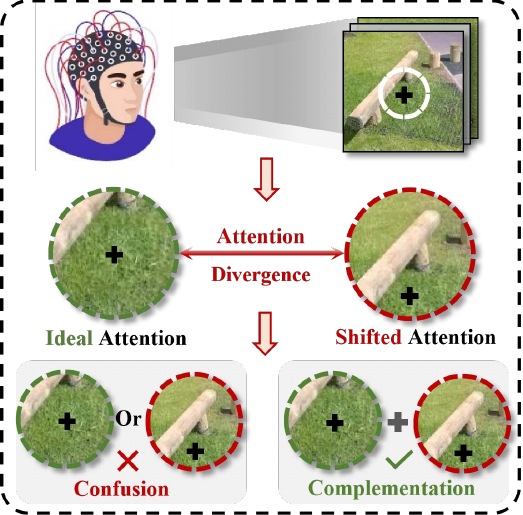}
    \caption{Attention drift.}
    \label{fig:motivation_a}
  \end{subfigure}
  \hfill
  \begin{subfigure}[t]{0.49\linewidth}
    \centering
    \includegraphics[width=\linewidth]{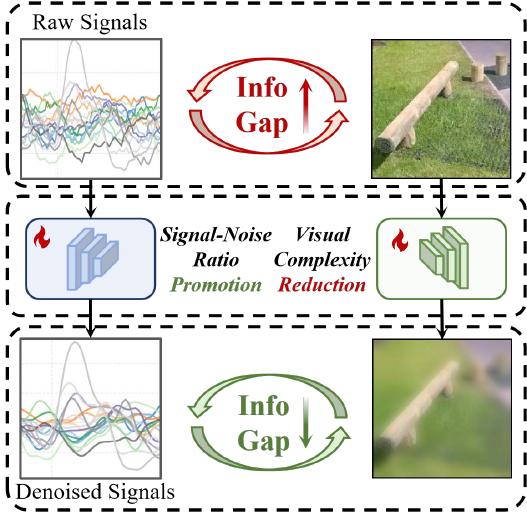}
    \caption{Bidirectional modulation.}
    \label{fig:motivation_b}
  \end{subfigure}
  \caption{Cognitive-Guided Visual–EEG Alignment. (a) Integrating center bias and saliency guidance to counteract attention drift. (b) Bidirectional Visual–EEG information alignment.}
  \label{motivation}
  \vspace{-4pt}
\end{wrapfigure}
Alternatively, some approaches concentrate on visual feature optimization. Bratrix \cite{t19} extracts image background and depth maps to emphasize structural information. BraVL \cite{t6} leverages high-level semantic embeddings from vision-language models to filter pixel-level redundancy. Nevertheless, these methods predominantly optimize unimodal representations without explicitly bridging the representational gap between neural signals and visual stimuli. Recently, UBP \cite{t20} proposes blurring images based on the center bias of human visual perception to alleviate this modality gap. However, this approach remains constrained by its rigid center-bias assumption, thereby neglecting attentional shifts triggered by peripheral salient targets. Furthermore, UBP lacks a concomitant mechanism to enhance EEG signal quality, hindering genuine bidirectional optimization and compromising robustness against individual variability and paradigm heterogeneity.

Human visual processing is non-uniform across the visual field. By default, fixations tend toward the image center due to high-resolution foveal vision \cite{t21,t22}. However, salient peripheral targets can capture attention during dynamic viewing \cite{t23,t24}. As shown in \cref{fig:motivation_a}, in typical EEG paradigms such as Rapid Serial Visual Presentation (RSVP), prolonged central fixation may induce attention fatigue and eye drift \cite{t25}. Consequently, a static center-blurring strategy could suppress neurally encoded peripheral information.

Moreover, EEG signals contain substantial task-irrelevant redundancy \cite{t26}. Given the distinct functional roles of neural oscillations across frequency bands \cite{t27}, such redundancy may reflect a functional mismatch between certain bands and the decoding task. This motivates frequency-band screening—retaining task-relevant bands while suppressing irrelevant ones—to systematically improve SNR. The information bottleneck theory offers a framework that seeks an optimal trade-off between the compression of an input variable and the preservation of relevant information about the downstream task \cite{t28}.

To address the above challenges, we propose \textbf{CAIA}, a Cognitive-guided Adaptive blurring with Information-Constrained Alignment framework for Neural-Visual decoding. As shown in \cref{fig:motivation_b}, CAIA bridges modality gaps via bidirectional information modulation. Visually, a cognitive-dynamics-based blurring mechanism combines center-fixation and saliency-guided paths, dynamically fused via EEG-driven cross-modal attention to simulate selective attention and reduce information density. On the EEG side, a cognitively guided information-screening method combines neural oscillation priors with an information-bottleneck-inspired spectral sparsity regularizer to select task-relevant frequency bands and suppress interference, enhancing SNR. A distribution-aware boundary calibration loss further identifies and calibrates outliers based on similarity statistics, improving robustness to attention shifts and noise. Our main contributions can be summarized as follows:
\begin{itemize}
\item \textbf{We propose a cognitive-dynamics-based brain-inspired visual decoding approach} that utilizes center fixation and saliency blurring, dynamically fusing the two visual paths via EEG-driven cross-modal attention to simulate visual selective attention and bridge the modality gap.
\item \textbf{We introduce a distribution-aware boundary calibration loss} to address outlier samples in EEG-image matching, which dynamically guides them toward the reference interval for robust cross-modal alignment.
\item \textbf{We develop a cognitively guided information-screening method} that integrates neuroscientific priors to automatically select task-relevant EEG oscillatory features, suppress irrelevant bands, and enhance SNR.
\item \textbf{Extensive experiments demonstrate that CAIA surpasses state-of-the-art methods}, achieving +19.4$\%$ Top-1 and +16.4$\%$ Top-5 accuracy gains over the strongest baseline UBP\cite{t20}.
\end{itemize}

\section{Related Works}
\subsection{Visual Neural Decoding}
Neural decoding aims to reconstruct perceptual and cognitive content from neural signals, with significant recent progress in motor imagery \cite{t29}, emotion recognition \cite{t30}, disease diagnosis \cite{t31}, and visual decoding \cite{t32,t33,t34,t8}. Visual decoding has garnered particular attention for its potential to elucidate the neural mechanisms underlying brain-environment interactions. While fMRI decoding benefits from high spatial resolution for detailed reconstruction \cite{t36,t37}, EEG decoding capitalizes on millisecond-scale temporal precision, rendering it particularly amenable to real-time brain-computer interfaces and the analysis of dynamic cognitive processes \cite{t12,t13}.

A core challenge in EEG visual decoding is establishing reliable cross-modal mappings. Early methods used handcrafted features that struggled with non-stationary trial dynamics \cite{t38}. Deep learning has since enabled specialized architectures \cite{t15,t16,t17,t39,t40,t41,t42} that improve EEG feature encoding. However, EEG's low SNR inherently limits encoder expressivity and mapping accuracy, and these methods do not address the fundamental inter-modal information granularity mismatch. Recent work has turned to visual feature optimization, via image decomposition or blurring \cite{t19,t20,t43}—to mitigate this granularity gap. However, these approaches fail to incorporate a neural feedback mechanism that coordinates dynamic visual adjustment with EEG SNR enhancement, limiting their resilience to shifts in cognitive states and unseen trials. In contrast, CAIA introduces a bidirectional co-optimization framework that combines EEG-conditioned adaptive visual fusion with neural frequency screening, thereby achieving more robust cross-modal alignment.

\subsection{Multi-Modal Contrastive Learning}
Contrastive learning encourages agreement between matched representations while distinguishing unmatched samples, showing strong results in vision \cite{t44} and language \cite{t45}. Models like CLIP leverage large-scale image-text pairs for zero-shot generalization \cite{t46}, inspiring its adaptation to neural decoding by aligning EEG representations with CLIP's visual space \cite{t47}. These methods typically use InfoNCE loss to pull matched pairs together and push unmatched pairs apart in a shared latent space.

Applying this paradigm directly to neural data faces two key issues. First, neural-visual pair quality is lower than natural image-text pairs. EEG contains substantial task-irrelevant noise (e.g., spontaneous oscillations, artifacts), leading to weak positive alignment and spurious correlations \cite{t48}. Second, common uniform optimization neglects variation in neural-visual pair similarities, whose distribution has been modeled using a Gaussian approximation in prior work \cite{t20}. Outliers (e.g., from attention drift or high noise) deviate from this main distribution, biasing gradients. Moreover, increasing model capacity or data augmentation may worsen noise overfitting \cite{t49}, underscoring the need to integrate cognitive mechanisms—not just data-driven learning. Our CAIA addresses these challenges by coupling EEG-guided visual adaptation with distribution-aware boundary calibration, mitigating cross-modal mismatches that may arise from attentional fluctuations and neural noise to promote more robust contrastive alignment.

\begin{figure*}[t]
    \centering
    \includegraphics[width=\textwidth]{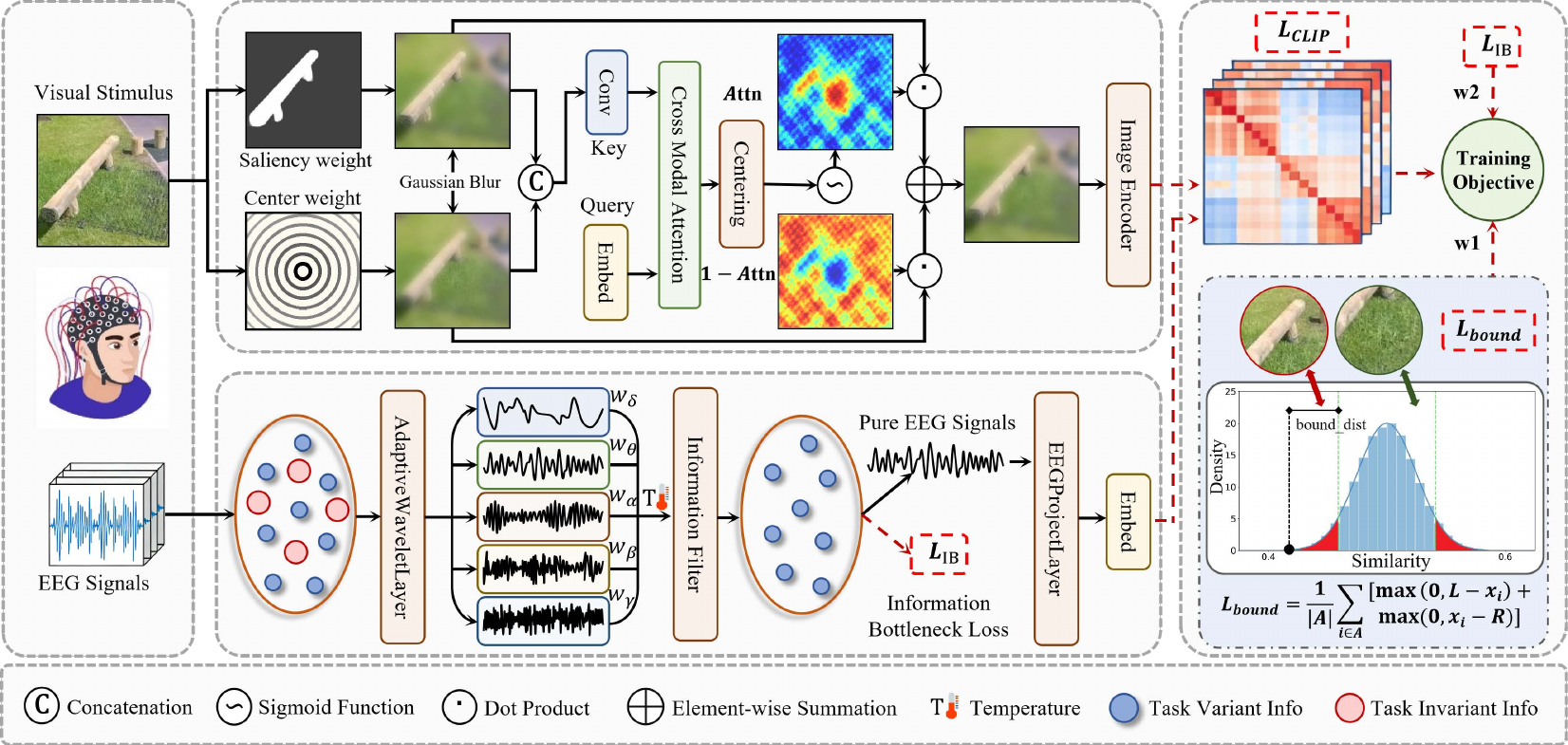}
    \caption{
    CAIA model framework: (a) dual-path adaptive blurring mechanism. (b) frequency-selective representation learning with IB-inspired spectral sparsity regularization. (c) distribution-aware boundary calibration loss.
    }
    \label{model}
\end{figure*}

\section{Method}
\label{sec:method}
\subsection{Overall Architecture}
CAIA introduces a cognitively guided framework that aligns EEG and visual representations through adaptive blurring and information constraints, as illustrated in \cref{model}. During training, EEG features are aligned with blurred image embeddings through contrastive learning, assisted by the boundary loss. During testing, EEG embeddings are matched to visual candidates based on similarity for retrieval evaluation.
\subsection{Cognitive-Dynamics-Based Adaptive Blurring Mechanism}
To mitigate information-density mismatch, we propose a dual-path adaptive blurring mechanism inspired by the complementary roles of center bias and stimulus-driven saliency in visual attention. Moving beyond static blurring strategies, this mechanism uses EEG features to adaptively balance the preservation of visual detail across image regions. It comprises two parallel blurring paths—saliency-guided differential blurring and center-guided radial blurring—which are fused via EEG-driven cross-modal attention to produce a neural-conditioned visual representation.
\subsubsection{Saliency-Guided Differential Blurring}
This branch leverages bottom-up visual saliency \cite{t24}. To simulate the “fine perception of salient regions, blurred perception of non-salient regions” characteristic, we apply spatially varying blur intensity inversely proportional to a saliency map. The saliency-guided blurred image $X_A$ is generated as:
\begin{align}
X_A(i,j) &= w_S(i,j) \cdot X(i,j) + \bigl[1 - w_S(i,j)\bigr] \cdot \mathcal{G}_\sigma(X)(i,j) \label{eq:Saliency_blur_main} \\
w_S(i,j) &= (1 - w_0) \cdot \text{saliency}(i,j) \label{eq:wS}
\end{align}
where $w_S(i,j)$ is the saliency-based preservation weight at pixel $(i,j)$, $w_0 \in [0,1]$ governs the overall strength of the blurring effect, $\text{saliency}(i,j)$ is obtained from a salient object detection model, and $\mathcal{G}_\sigma$ is a Gaussian blur operator.

\subsubsection{Center-Guided Radial Blurring}
This path incorporates the prior of visual center bias—the tendency of the human visual system to direct initial and early fixations toward the geometric center of stimuli \cite{t22}. We simulate this characteristic using a radial blur profile that preserves clarity at the center and gradually increases blur toward the periphery \cite{t20}. The center-guided blurred image $X_B$ is computed as:
\begin{align}
X_B(i,j) &= w_R(i,j) \cdot X(i,j) + \bigl[1 - w_R(i,j)\bigr] \cdot \mathcal{G}_\sigma(X)(i,j)\\
w_R(i,j) &= (1 - w_0) \cdot \exp\bigl(-g \cdot d(i,j)\bigr)
\label{eq:center_blur}
\end{align}
where $w_R(i,j)$ is the center-bias preservation weight at pixel $(i,j)$, $w_0 \in [0,1]$ governs the overall strength of the blurring effect, $d(i,j)$ is the normalized distance from the image center, and $g$ controls the blur decay rate.

\subsubsection{Cross-Modal Attention Fusion Strategy}
The two blurring paths have complementary strengths and context-dependent applicability. To dynamically balance the “default center” and “stimulus capture” mechanisms, we introduce a cross-modal attention fusion strategy. Here, EEG features serve as the query $Q$, providing neural context for adaptive fusion, while features from the concatenated dual-path blurred images form the key $K$. This allows the model to estimate pixel-wise fusion weights $A$ that are informed by the neural signal:
\begin{align}
S &= Q^\top K \in \mathbb{R}^{1 \times HW} \\
A &= \operatorname{sigmoid}\bigl(S - \bar{S}\bigr) \in \mathbb{R}^{1 \times HW} \\
X_{\text{fused}} &= A \odot X_A + (1-A) \odot X_B
\label{eq:attention_fusion}
\end{align}
We adopt the sigmoid activation in lieu of softmax to compute pixel-wise fusion weights. This design avoids the sum-to-one constraint imposed by softmax, which can cause overconfidence in a single region and limit attention allocation across multiple salient areas \cite{t55}. The sigmoid function enables independent modulation per pixel, permitting simultaneous attention to several regions. To preserve relative spatial contrast amid this independence, we apply a centering operation by subtracting the mean score $\bar{S}$, thereby amplifying differential attention strength across the image and enhancing the spatial coherence of the fused output.

\subsection{Cognitively Guided Information-Screening Method}
This module selects task-relevant spectral information from EEG signals by leveraging neuroscientific priors on the frequency-specific functional roles of neural oscillations. Its design is inspired by the Information Bottleneck (IB) principle, which emphasizes preserving task-relevant information while limiting irrelevant input information \cite{t28}.

In visual decoding, the EEG representation should preserve stimulus-related features while reducing sensitivity to irrelevant spectral activity. Cross-modal contrastive learning guides the representation toward the visual stimulus, and a regularized spectral selection stage controls how different frequency components contribute to EEG encoding.

Specifically, we first decompose EEG signals into sub-bands using an adaptive wavelet transform. The decomposition scale $k_i=k_i^0\cdot\gamma_i$ is adjusted through a learnable parameter $\gamma_i$, enabling task-specific spectral tuning. A spectral gating layer then assigns soft selection weights $\{m_i\}$ to the sub-bands through a temperature-controlled softmax:
\begin{equation}
m_i = \frac{\exp(w_i / \tau)}{\sum_{j=1}^{n} \exp(w_j / \tau)}
\label{eq:softmax_temperature}
\end{equation}
where $w_i$ are learnable logits, $n$ is the number of sub-bands, and $\tau$ controls the concentration of the selection weights. The gated sub-band features are fused and passed to the EEG encoder to obtain $Z$.

To encourage selective spectral usage, we regularize the entropy of the band-selection distribution $M=\{m_i\}_{i=1}^{n}$:
\begin{equation}
H(M) = -\sum_{i=1}^{n} m_i \log m_i
\label{eq:spectral_entropy}
\end{equation}
Minimizing $H(M)$ encourages the weights to concentrate on fewer sub-bands. The scales and band weights are learned jointly with the contrastive objective, so the emphasis on each spectral component evolves with its contribution to neural-visual alignment.

\subsection{Distribution-Aware Boundary Calibration Loss}
\begin{wrapfigure}{r}{0.48\textwidth}
  \vspace{-10pt}
  \centering
  \includegraphics[width=\linewidth]{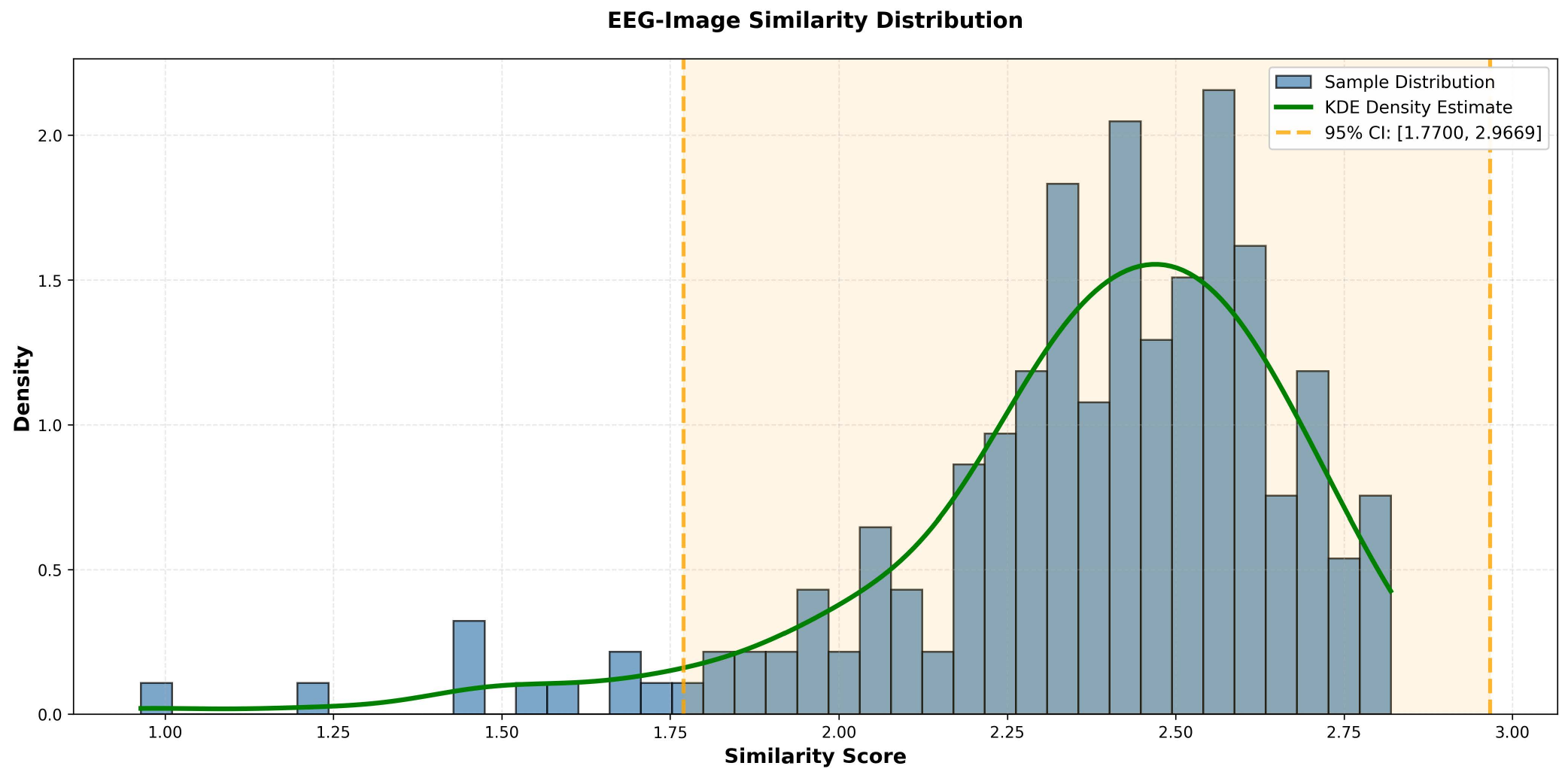}
  \caption{EEG-image similarity distribution.}
  \label{orign_distribution}
  \vspace{-8pt}
\end{wrapfigure}

Matched EEG-visual pairs exhibit varying degrees of alignment. Attentional fluctuations and neural noise may contribute to atypical similarity scores, introducing cross-modal mismatches that can interfere with contrastive learning. We therefore use the similarity distribution to construct an adaptive reference interval for boundary calibration. A Gaussian reference model provides a location-and-scale approximation for defining this interval, following prior work \cite{t20}. The observed similarity distribution is illustrated in \cref{orign_distribution}.

Let $s_i$ denote the similarity score of the $i$-th matched pair, and let $\bar{s}$ and $\sigma$ be running estimates of the mean and standard deviation, updated from training-batch similarity statistics. With $z_{1-\alpha/2}=\Phi^{-1}(1-\alpha/2)$, where $\Phi$ is the standard normal cumulative distribution function, the reference interval is $[\bar{s}-z_{1-\alpha/2}\sigma,\;\bar{s}+z_{1-\alpha/2}\sigma]$. Matched pairs outside this interval form the set $\mathcal{M}$ and are penalized according to their distance from the corresponding boundary:
\begin{align}
\mathcal{L}_{\text{bound}} &= \frac{1}{|\mathcal{M}|} \sum_{i \in \mathcal{M}} \bigl(\ell_i^{\text{left}} + \ell_i^{\text{right}}\bigr) \\
\ell_i^{\text{left}} &= \max\bigl(\bar{s} - z_{1-\alpha/2}\sigma - s_i, 0\bigr) \\
\ell_i^{\text{right}} &= \max\bigl(s_i - \bar{s} - z_{1-\alpha/2}\sigma, 0\bigr)
\label{eq:boundary_loss}
\end{align}
When $\mathcal{M}$ is empty, we define $\mathcal{L}_{\text{bound}}=0$.

The penalty draws outlying scores toward the reference interval, moderating extreme deviations during contrastive alignment. Its gradients jointly adjust the EEG and fused visual representations, while the running statistics allow the reference boundaries to follow the evolving similarity distribution. Per-subject distributional diagnostics and comparisons with empirical percentile boundaries are provided in Appendix~\ref{app:boundary_calibration}.

Combining the components, our full training objective is:
\begin{equation}
\mathcal{L}_{\text{Overall}} = \mathcal{L}_{\text{CLIP}} + w_1\mathcal{L}_{\text{bound}} + w_2 H(M)
\label{eq:overall_loss}
\end{equation}
where $\mathcal{L}_{\text{CLIP}}$ is the CLIP-based contrastive loss \cite{t46}, $H(M)$ is the entropy of the band-selection distribution, and $w_1$ and $w_2$ balance boundary calibration and spectral regularization.

\section{Experiment}
\label{sec:experiments}
\subsection{Datasets and Implementation Details}
\label{sec:implementation}
We evaluate the proposed framework and comparison methods on two publicly available neuroimaging datasets derived from the THINGS stimulus library.

THINGS-EEG \cite{t14} records electroencephalography (EEG) from 10 subjects under an RSVP paradigm. The training set comprises 1,654 categories (10 images per category, 4 repetitions each), while the test set contains 200 novel categories (1 image per category, 80 repetitions). Following UBP \cite{t20}, we enhance the signal-to-noise ratio by averaging repeated trials of the same stimulus, using a post-stimulus window of 0–250 ms and selecting 17 visually related channels spanning occipital, parietal, and lateral temporal regions.

THINGS-MEG \cite{t52} contains magnetoencephalography (MEG) recordings from 4 subjects (271 channels). The training set covers all 1,654 categories (12 images per category, single presentation), and the test set comprises 200 novel categories (1 image per category, 12 repetitions).

We adopt the ResNet-50 vision encoder from the OpenCLIP library as our visual feature extractor. All experiments were implemented in PyTorch on an NVIDIA RTX 3090 GPU. We used the Adam optimizer with a batch size of 128 and a learning rate of 1e-4 for subject-dependent settings, and a batch size of 256 with a learning rate of 1e-5 for subject-independent settings. The loss weights $w_1$ and $w_2$ in \cref{eq:overall_loss} are set to 0.01. The blur decay rate $g$ is set to 3.

\subsection{Overall Performance}
\begin{table*}[t]
  \centering
  \Huge
  \caption{Overall accuracy (\%) of 200-way zero-shot retrieval on THINGS-EEG: Top-1 and Top-5.}
\resizebox{\linewidth}{!}{
  \begin{tabular}{lcccccccccccccccccccccc}
    \toprule  
    & \multicolumn{2}{c}{Subject 1} & \multicolumn{2}{c}{Subject 2} & \multicolumn{2}{c}{Subject 3} & \multicolumn{2}{c}{Subject 4} & \multicolumn{2}{c}{Subject 5} & \multicolumn{2}{c}{Subject 6} & \multicolumn{2}{c}{Subject 7} & \multicolumn{2}{c}{Subject 8} & \multicolumn{2}{c}{Subject 9} & \multicolumn{2}{c}{Subject 10} & \multicolumn{2}{c}{Avg} \\
    \cmidrule(r){2-3} \cmidrule(r){4-5} \cmidrule(r){6-7} \cmidrule(r){8-9} \cmidrule(r){10-11} \cmidrule(r){12-13} \cmidrule(r){14-15} \cmidrule(r){16-17} \cmidrule(r){18-19} \cmidrule(r){20-21} \cmidrule(r){22-23}
    Method & Top-1 & Top-5 & Top-1 & Top-5 & Top-1 & Top-5 & Top-1 & Top-5 & Top-1 & Top-5 & Top-1 & Top-5 & Top-1 & Top-5 & Top-1 & Top-5 & Top-1 & Top-5 & Top-1 & Top-5 & Top-1 & Top-5 \\
    \midrule
    \specialrule{0em}{1.5pt}{1.5pt}
    \midrule
    \multicolumn{23}{c}{Subject-dependent (intra-subject): train and test on one subject} \\
    \midrule
    BraVL~ \cite{t6} & 6.1 & 17.9 & 4.9 & 14.9 & 5.6 & 17.4 & 5.0 & 15.1 & 4.0 & 13.4 & 6.0 & 18.2 & 6.5 & 20.4 & 8.8 & 23.7 & 4.3 & 14.0 & 7.0 & 19.7 & 5.8 & 17.5 \\    
    NICE~ \cite{t16} & 13.2 & 39.5 & 13.5 & 40.3 & 14.5 & 42.7 & 20.6 & 52.7 & 10.1 & 31.5 & 16.5 & 44.0 & 17.0 & 42.1 & 22.9 & 56.1 & 15.4 & 41.6 & 17.4 & 45.8 & 16.1 & 43.6 \\
    ATM~ \cite{t17} & 25.6 & 60.4 & 22.0 & 54.5 & 25.0 & 62.4 & 31.4 & 60.9 & 12.9 & 43.0 & 21.3 & 51.1 & 30.5 & 61.5 & 38.8 & 72.0 & 24.4 & 51.5 & 29.1 & 63.5 & 26.1 & 58.1 \\
    CogCap~ \cite{t43} & 27.2 & 59.5 & 28.7 & 57.0 & 37.2 & 66.1 & 37.7 & 63.2 & 21.8 & 47.8 & 31.6 & 58.1 & 32.8 & 59.6 & 47.6 & 73.5 & 33.4 & 57.7 & 35.1 & 63.6 & 33.3 & 60.6 \\
    VE-SDN~ \cite{t47} & 32.6 & 63.7 & 34.4 & 69.9 & 38.7 & 73.5 & 39.8 & 72.0 & 29.4 & 58.6 & 34.5 & 68.8 & 34.5 & 68.3 & 49.3 & 79.8 & 39.0 & 69.6 & 39.8 & 75.3 & 37.2 & 70.0 \\
    MB2C~ \cite{t51} & 23.6 & 56.3 & 22.6 & 50.5 & 26.3 & 60.1 & 34.8 & 67.0 & 21.3 & 53.0 & 31.0 & 62.3 & 25.0 & 54.8 & 39.0 & 69.3 & 27.5 & 59.3 & 33.1 & 70.8 & 28.4 & 60.3 \\
    Neural-MCRL~ \cite{t50} & 27.5 & 64.0 & 28.5 & 61.5 & 37.0 & 69.0 & 35.0 & 66.0 & 22.5 & 51.5 & 31.5 & 61.0 & 31.5 & 62.5 & 42.0 & 74.5 & 30.5 & 59.5 & 37.5 & 71.0 & 32.4 & 64.1 \\
    UBP~ \cite{t20} & \underline{41.2} & \underline{70.5} & \underline{51.2} & \underline{80.9} & \underline{51.2} & \underline{82.0} & \textbf{51.1} & \underline{76.9} & \underline{42.2} & \underline{72.8} & \underline{57.5} & \underline{83.5} & \underline{49.0} & \underline{79.9} & \underline{58.6} & \underline{85.8} & \underline{45.1} & \underline{76.2} & \underline{61.5} & \underline{88.2} & \underline{50.9} & \underline{79.7} \\
    \rowcolor{blue!10} 
    \textbf{CAIA (Ours)} & \textbf{74.0} & \textbf{98.0} & \textbf{80.0} & \textbf{97.5} & \textbf{69.0} & \textbf{94.0} & \underline{51.0} & \textbf{92.5} & \textbf{67.5} & \textbf{94.0} & \textbf{67.5} & \textbf{96.5} & \textbf{63.0} & \textbf{95.5} & \textbf{83.0} & \textbf{99.0} & \textbf{69.0} & \textbf{96.5} & \textbf{79.0} & \textbf{97.5} & \textbf{70.3} & \textbf{96.1} \\
    \midrule
    \specialrule{0em}{1.5pt}{1.5pt}
    \midrule
    \multicolumn{23}{c}{Subject-independent (inter-subject): leave one subject out for test} \\
    \midrule
    BraVL~ \cite{t6} & 2.3 & 8.0 & 1.5 & 6.3 & 1.4 & 5.9 & 1.7 & 6.7 & 1.5 & 5.6 & 1.8 & 7.2 & 2.1 & 8.1 & 2.2 & 7.6 & 1.6 & 6.4 & 2.3 & 8.5 & 1.8 & 7.0 \\  
    NICE~ \cite{t16} & 7.6 & 22.8 & 5.9 & 20.5 & 6.0 & 22.3 & 6.3 & 20.7 & 4.4 & 18.3 & 5.6 & 22.2 & 5.6 & 19.7 & 6.3 & 22.0 & 5.7 & 17.6 & 8.4 & 28.3 & 6.2 & 21.4 \\
    ATM~ \cite{t17} & 10.5 & 26.8 & 7.1 & 24.8 & 11.9 & 33.8 & \underline{14.7} & \underline{39.4} & 7.0 & 23.9 & 11.1 & \underline{35.8} & \underline{16.1} & \textbf{43.5} & 15.0 & \underline{40.3} & 4.9 & 22.7 & \underline{20.5} & \underline{46.5} & 11.9 & 33.8 \\
    CogCap~ \cite{t43} & \textbf{16.3} & \textbf{42.3} & \underline{16.2} & 37.9 & 8.8 & 26.8 & \textbf{15.4} & 37.6 & 10.1 & 31.7 & \underline{14.0} & 35.4 & 10.7 & 26.9 & 13.9 & 34.2 & 9.0 & 32.4 & 15.3 & 38.6 & 13.0 & \underline{34.4} \\
    MB2C~ \cite{t51} & 10.5 & 28.1 & 11.3 & 32.8 & 8.8 & 27.6 & 13.6 & 33.5 & 10.6 & 27.5 & 12.1 & 33.1 & 11.5 & 31.8 & 12.0 & 32.1 & 12.1 & 31.3 & 16.1 & 42.1 & 11.9 & 32.0 \\
    Neural-MCRL~ \cite{t50} & 13.0 & 31.5 & 12.0 & 30.5 & \textbf{14.5} & \underline{35.5} & 12.5 & 35.5 & \underline{11.5} & 29.0 & 13.5 & 35.5 & 14.0 & 36.0 & \textbf{18.5} & 38.5 & \underline{13.5} & 32.5 & 17.0 & 39.0 & \underline{14.0} & 34.3 \\
    UBP~ \cite{t20} & 11.5 & 29.7 & 15.5 & \underline{40.0} & 9.8 & 27.0 & 13.0 & 32.3 & 8.8 & \underline{33.8} & 11.7 & 31.0 & 10.2 & 23.8 & 12.2 & 32.2 & \textbf{15.5} & \textbf{40.5} & 16.0 & 43.5 & 12.4 & 33.4 \\
    \rowcolor{blue!10} 
    \textbf{CAIA (Ours)} & \underline{14.0} & \underline{39.5} & \textbf{21.5} & \textbf{57.5} & \underline{13.5} & \textbf{37.0} & 14.0 & \textbf{53.5} & \textbf{12.0} & \textbf{40.5} & \textbf{16.0} & \textbf{43.0} & \textbf{16.5} & \underline{42.0} & \underline{15.0} & \textbf{40.5} & 10.0 & \underline{35.5} & \textbf{31.0} & \textbf{67.5} & \textbf{16.4} & \textbf{45.7} \\
    \bottomrule
  \end{tabular}} 
  \vspace{0.1cm}
  \vspace{-0.2cm} 
  \label{tab:main_results}
\end{table*}

We compare the proposed method with state-of-the-art approaches, including BraVL \cite{t6}, NICE \cite{t16}, ATM \cite{t17}, CogCap \cite{t43}, VE-SDN \cite{t47}, MB2C \cite{t51}, Neural-MCRL \cite{t50}, and UBP \cite{t20}.

\begin{wraptable}{r}{0.5\textwidth}
  \centering
  \vspace{-8pt}
  \caption{Overall Top-1/5 accuracy (\%) on THINGS-MEG for 200-way zero-shot recognition.}
  \resizebox{0.5\textwidth}{!}{%
  \renewcommand{\arraystretch}{1.0}
  \setlength{\tabcolsep}{3.5pt}
  \begin{tabular}{lcccccccccc}
    \toprule
    & \multicolumn{2}{c}{Sub 1} & \multicolumn{2}{c}{Sub 2} & \multicolumn{2}{c}{Sub 3} & \multicolumn{2}{c}{Sub 4} & \multicolumn{2}{c}{Avg} \\
    \cmidrule(r){2-3} \cmidrule(r){4-5} \cmidrule(r){6-7} \cmidrule(r){8-9} \cmidrule(r){10-11}
    Method & T1 & T5 & T1 & T5 & T1 & T5 & T1 & T5 & T1 & T5 \\
    \midrule
    \multicolumn{11}{c}{Subject-dependent} \\
    \midrule
    MB2C~\cite{t51} & 9.3 & 33.6 & 20.6 & 49.2 & 18.2 & 44.3 & 10.2 & 33.6 & 14.6 & 40.2 \\
    NICE~\cite{t16} & 11.5 & 35.6 & 25.7 & 54.4 & 21.0 & 47.8 & 11.2 & 35.2 & 17.4 & 43.3 \\
    ATM~\cite{t17} & 8.0 & 29.3 & 30.2 & 61.5 & 20.3 & 50.5 & 11.8 & 33.3 & 17.6 & 43.7 \\
    UBP~\cite{t20} & \underline{15.0} & \textbf{38.0} & \underline{46.0} & \underline{80.5} & \underline{27.3} & \underline{59.0} & \underline{18.5} & \underline{43.5} & \underline{26.7} & \underline{55.2} \\
    \rowcolor{blue!10}
    \textbf{CAIA} & \textbf{18.5} & \textbf{38.0} & \textbf{47.0} & \textbf{89.0} & \textbf{37.0} & \textbf{73.5} & \textbf{24.5} & \textbf{50.0} & \textbf{31.8} & \textbf{62.6} \\
    \midrule
    \multicolumn{11}{c}{Subject-independent} \\
    \midrule
    UBP~\cite{t20} & 2.0 & 5.7 & 1.5 & \textbf{17.2} & 2.7 & 10.5 & 2.5 & 8.0 & 2.2 & 10.4 \\
    \rowcolor{blue!10}
    \textbf{CAIA} & \textbf{9.5} & \textbf{27.0} & \textbf{4.0} & 15.0 & \textbf{9.5} & \textbf{21.5} & \textbf{7.0} & \textbf{23.5} & \textbf{7.5} & \textbf{21.8} \\
    \bottomrule
  \end{tabular}%
  }
  \label{tab:meg}
  \vspace{-8pt}
\end{wraptable}
\vspace{0pt}

As summarized in \cref{tab:main_results} and \cref{tab:meg}, CAIA establishes a new state of the art on both EEG and MEG test sets under subject-dependent and subject-independent protocols.

On THINGS-EEG, CAIA attains subject-dependent Top‑1/Top‑5 accuracy of \textbf{70.3$\%$ / 96.1$\%$}, surpassing the UBP baseline by +19.4$\%$ / +16.4$\%$. This marked improvement validates that the proposed adaptive blurring mechanism and EEG‑side information screening collectively mitigate the cross‑modal granularity mismatch. In the more challenging subject‑independent setting, CAIA achieves \textbf{16.4$\%$ / 45.7$\%$} (UBP: 12.4$\%$ / 33.4$\%$), indicating improved robustness to inter-subject variability.

On THINGS-MEG, CAIA achieves subject-dependent accuracies of \textbf{31.8$\%$ / 62.6$\%$} and subject-independent accuracies of \textbf{7.5$\%$ / 21.8$\%$}, outperforming UBP by +5.1$\%$ / +7.4$\%$ and +5.3$\%$ / +11.4$\%$, respectively. These results indicate that the framework can leverage neural oscillation priors to identify stable cross-modal relationships even with weaker inter-subject signals, evidencing promising cross-modal generalization.

\subsection{Ablation Experiments}
We conduct ablation studies to evaluate three core components of CAIA individually and in combination: the adaptive blurring mechanism ($ATTN$), the information-screening representation learner ($IBWave$), and the boundary calibration loss ($L_{bound}$). Results are reported in \cref{tab:ablation}.

Disabling all modules (leaving only the EEG projection head, center‑biased blurring, and the CLIP contrastive loss) yields a baseline performance of 54.5$\%$ Top‑1 and 87.3$\%$ Top‑5 accuracy on THINGS‑EEG. Enabling $ATTN$ alone improves performance to 62.1$\%$ / 92.6$\%$, demonstrating the efficacy of center-saliency fusion in reducing visual redundancy. Using $IBWave$ in isolation achieves 64.0$\%$ / 91.6$\%$, indicating that frequency-band screening suppresses EEG noise and enhances the SNR. Employing $L_{bound}$ alone leads to a slight degradation (52.4$\%$ / 85.7$\%$). This may reflect the difficulty of imposing a similarity-range constraint on representations that still contain substantial cross-modal mismatch. By identifying atypical pair similarities and guiding them toward the reference interval, the boundary loss provides a calibration signal for cross-modal alignment. Combined with adaptive visual fusion and spectral screening, this signal guides representation refinement through adjustments to visual detail and neural spectral emphasis.

\begin{wraptable}{r}{0.48\textwidth}
  \centering
  \caption{Ablation study on THINGS-EEG (average Top‑1~/~Top‑5, \%).}
  \label{tab:ablation}
  \begin{tabular}{c@{\hspace{4pt}}c@{\hspace{4pt}}c@{\quad}c@{\hspace{4pt}}c}
    \toprule
    \multicolumn{3}{c}{Modules} & \multicolumn{2}{c}{Avg}\\
    \cmidrule(r){1-3}\cmidrule(l){4-5}
    IBWave & ATTN & $L_{\text{bound}}$ & Top-1 & Top-5\\
    \midrule
    $\times$ & $\times$ & $\times$ & 54.5 & 87.3\\
    $\times$ & $\times$ & \checkmark & 52.4 & 85.7\\
    $\times$ & \checkmark & $\times$ & 62.1 & 92.6\\
    $\times$ & \checkmark & \checkmark & 64.0 & 93.0\\
    \checkmark & $\times$ & $\times$ & 64.0 & 91.6\\
    \checkmark & $\times$ & \checkmark & 64.9 & 92.4\\
    \checkmark & \checkmark & $\times$ & 64.9 & 93.6\\
    \rowcolor{blue!10}
    \checkmark & \checkmark & \checkmark & \textbf{70.3} & \textbf{96.1}\\
    \bottomrule
  \end{tabular}
  \vspace{-6pt}
\end{wraptable}
\vspace{0pt}

Pairwise combinations reveal synergistic effects. $ATTN+IBWave$ attains 64.9$\%$ / 93.6$\%$, highlighting complementary benefits from visual and EEG-side optimization. $ATTN+L_{bound}$ yields 64.0$\%$ / 93.0$\%$, supporting the complementary roles of adaptive fusion and boundary calibration. $IBWave+L_{bound}$ achieves 64.9$\%$ / 92.4$\%$, as noise-suppressed EEG features facilitate more reliable distribution modeling.

The full CAIA model integrates all three components, achieving the best overall performance of 70.3$\%$ Top-1 and 96.1$\%$ Top-5 accuracy. This optimal result demonstrates the effectiveness of the complete design loop: $IBWave$ supplies denoised EEG features, $ATTN$ generates cognitively aligned blurred images, and $L_{bound}$ calibrates residual mismatches, collectively establishing robust cross-modal alignment.

\subsection{Model Analysis}
\subsubsection{Cross-Modal Attention Analysis}
\begin{wrapfigure}{r}{0.48\textwidth}
  \centering
  \vspace{-12pt}
  \includegraphics[width=\linewidth]{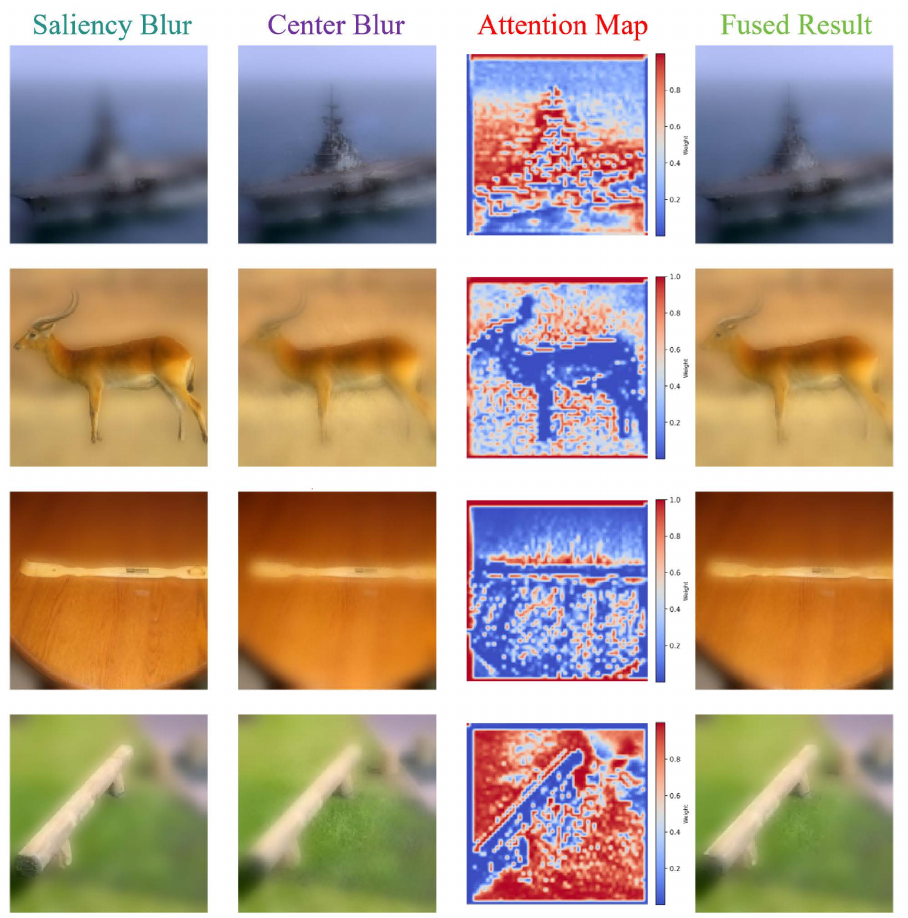}
  \caption{
  Visualization of the cognitive-dynamics-based adaptive blurring mechanism. Red regions indicate greater weighting of the center-biased pathway, whereas blue regions indicate greater weighting of the saliency-guided pathway.
  }
  \label{attention}
  \vspace{-28pt}
\end{wrapfigure}

As shown in \cref{attention}, the adaptive blurring mechanism produces spatially varying fusion weights, balancing center-biased and saliency-guided processing across image regions. This allows visual detail to be selectively preserved through complementary blurring pathways.

To evaluate the effectiveness of cross-modal attention, we compare the proposed mechanism with alternative image-processing strategies, including the original image, center-only blur, saliency-only blur, fixed linear blending, and a global learnable mask. Quantitative results in \cref{tab:image_fusion} show that EEG-guided fusion outperforms these alternatives. By conditioning the balance between center-bias and saliency cues on neural features, the mechanism adapts visual detail to the neural context, supporting more effective cross-modal alignment.

\subsubsection{Information-Guided Frequency Band Screening Analysis}
After information-guided screening, the Beta band receives the highest selection weight (see Appendix~\ref{app:frequency_screening} for details), consistent with its reported involvement in visual attention, feature integration, and target recognition. Beta oscillations have been associated with information flow between visual and frontoparietal attention networks, providing a neurophysiological context for the learned spectral preference \cite{t53,t54}.

\begin{figure*}[t]
    \centering
    \includegraphics[width=\textwidth]{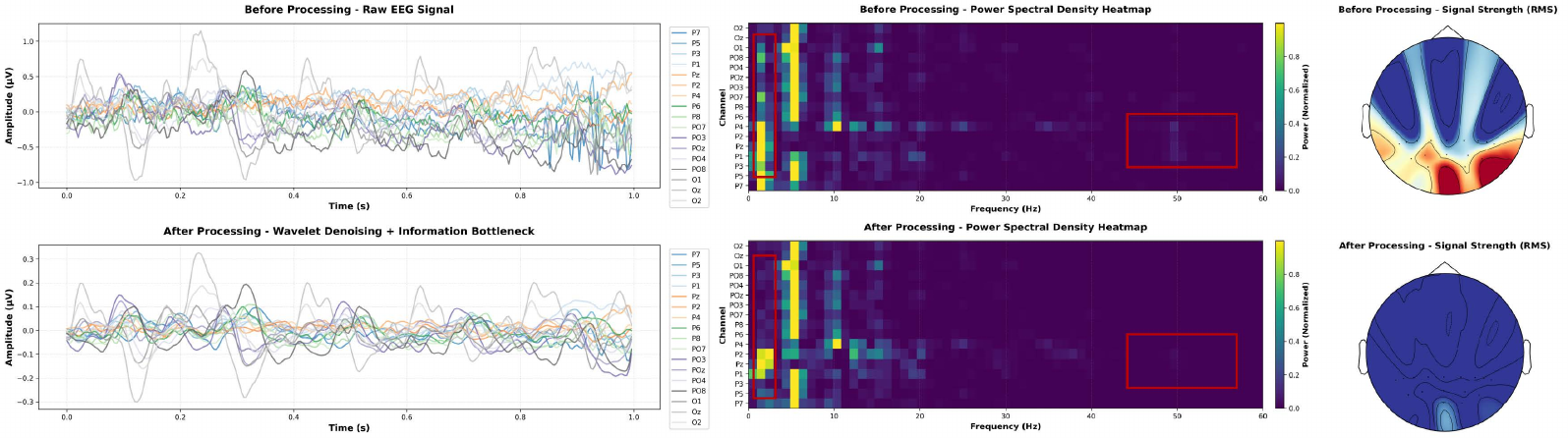}
    \caption{
    Visualization of EEG signals before and after information-guided frequency band screening in the time domain (left), frequency domain (middle), and EEG topographic map (right).
    }
    \label{denoise}
\end{figure*}

Time-frequency visualizations illustrate the spectral reshaping induced by the IB-inspired screening module (\cref{denoise}). The signal amplitude decreases from approximately $[-1, 1]$ to $[-0.3, 0.3]$, accompanied by selective attenuation across frequency ranges. Components above 40 Hz, which may include EMG-related interference, are markedly attenuated, while activity below 5 Hz is reduced in certain channels. These patterns are consistent with the module's denoising motivation, reducing potentially interfering spectral activity while emphasizing components relevant to visual decoding.

\begin{figure}[ht]
  \centering
  \begin{subfigure}{0.48\textwidth}
    \centering
    \includegraphics[width=\linewidth]{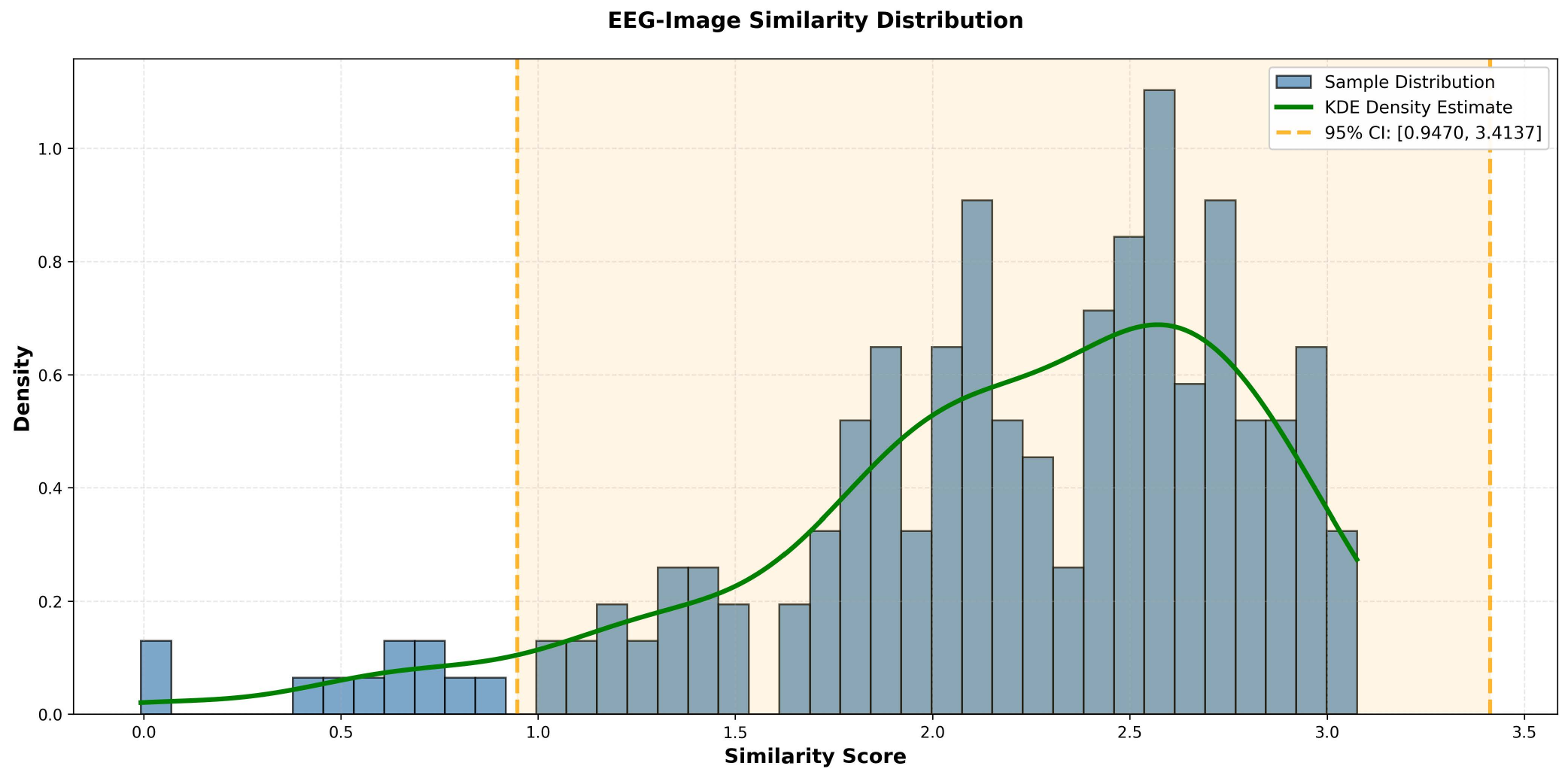}
    \caption{before}
    \label{fig:a}
  \end{subfigure}
  \hfill  
  \begin{subfigure}{0.48\textwidth}
    \centering
    \includegraphics[width=\linewidth]{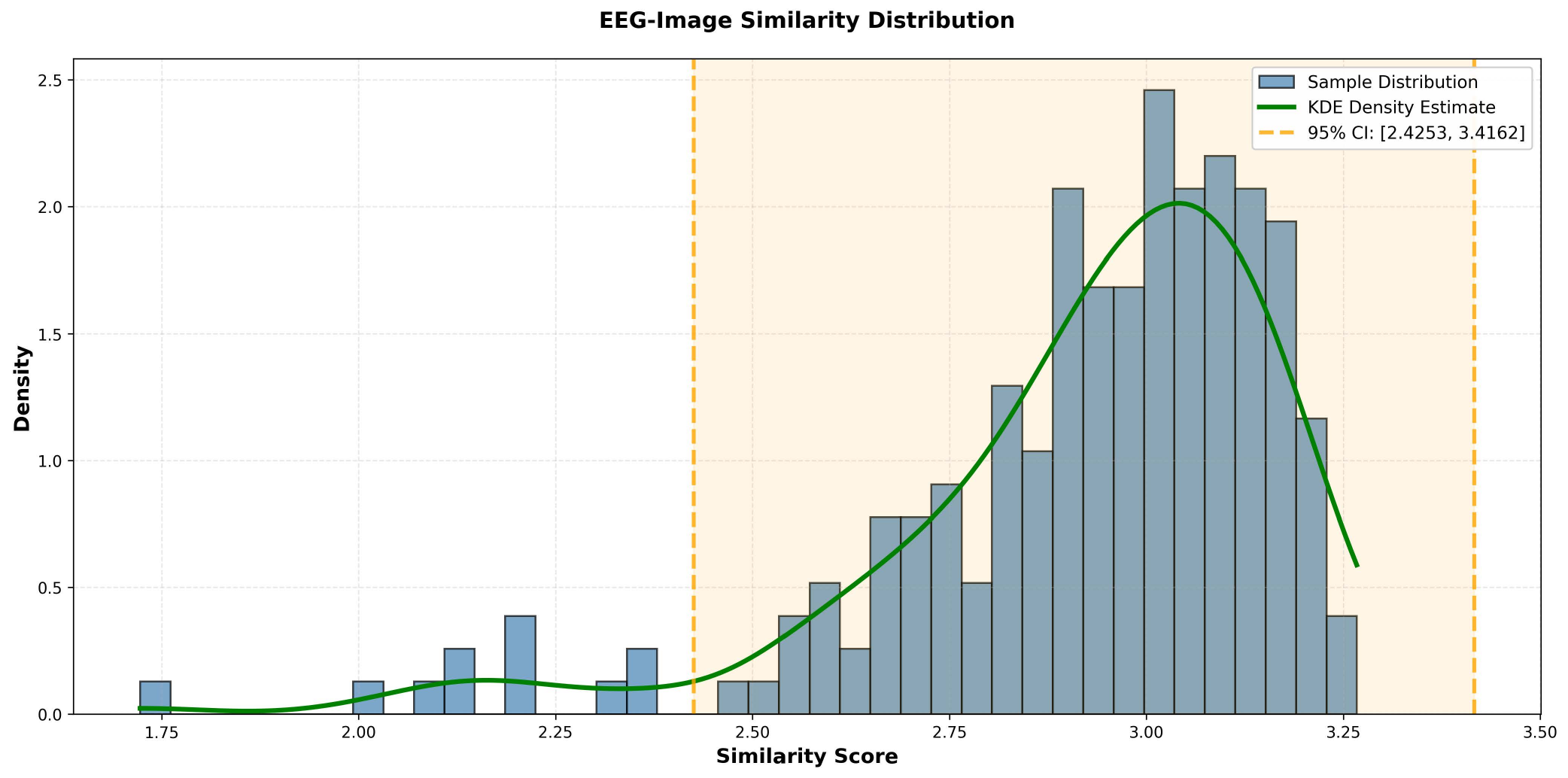}
    \caption{after}
    \label{fig:b}
  \end{subfigure}
  \caption{EEG-image similarity distribution before and after calibration using the distribution-aware loss.}
  \label{fig:distribution}
\end{figure}

The EEG topography (\cref{denoise}) shows a spatial concentration of signal intensity over the occipital-parietal region, consistent with prior reports \cite{t26}. This posterior pattern may reflect contributions from visual processing in early visual areas (V1–V4) and attentional modulation in parietal regions, including those around the intraparietal sulcus, offering a neurophysiological interpretation of the spatial emphasis learned by the screening module.

\subsubsection{Calibration Effect of Distribution-Aware Boundary Loss}
\begin{wraptable}{r}{0.48\textwidth}
  \centering
  \vspace{-12pt}
  \renewcommand{\arraystretch}{1.0}
  \caption{Image-fusion comparison on THINGS-EEG (average Top-1 / Top-5, \%).}
  \begin{tabular}{lcc}
    \toprule
    Method & Top-1 & Top-5 \\
    \midrule
    Original          & 64.6 & 92.1 \\
    Center\_blur      & \underline{64.9} & \underline{92.4} \\
    Saliency\_blur    & 59.3 & 88.0 \\
    Linear            & 63.2 & 91.5 \\
    Learnable\_mask   & 64.1 & 92.1 \\
    \rowcolor{blue!10} \textbf{Attention (Ours)} & \textbf{70.3} & \textbf{96.1} \\
    \bottomrule
  \end{tabular}
  \label{tab:image_fusion}
\end{wraptable}
As depicted in \cref{fig:distribution}, incorporating the distribution-aware boundary loss into CAIA yields a more compact EEG-image similarity distribution, with a higher mean similarity and fewer outliers beyond the reference interval. This pattern is consistent with its calibration role in guiding atypical pair similarities toward the reference interval while contrastive learning strengthens cross-modal alignment. By moderating extreme deviations that may arise from attentional fluctuations or neural noise, boundary calibration complements adaptive visual fusion and spectral screening in refining neural-visual alignment.

\section{Conclusion}
\label{sec:conclusion}
CAIA presents a cognitively guided framework that mitigates the information granularity mismatch in EEG‑visual decoding via bidirectional modulation. Visually, an adaptive blurring mechanism fuses center‑biased and saliency‑guided cues through EEG‑driven cross‑modal attention. On the neural side, an IB-inspired frequency-screening module selectively enhances task-relevant frequency bands to improve SNR, while a distribution-aware boundary loss calibrates alignment outliers.

Extensive experiments on THINGS‑EEG and THINGS‑MEG demonstrate that CAIA consistently surpasses state‑of‑the‑art methods in both subject‑dependent and subject‑independent zero‑shot retrieval. Ablation studies support the complementary roles of the proposed components, while visual analyses illustrate adaptive visual fusion and selective attenuation of potentially interfering spectral activity.

The adopted cognitive priors—center bias and saliency capture—presuppose an experimental paradigm with sustained central fixation, and their effectiveness may degrade under free-viewing conditions or paradigms without a well-defined fixation point. The current evaluation also relies on repeated-trial averaging and does not incorporate eye-tracking-based validation of fixation patterns or attentional shifts. This work offers a principled, cognition-informed approach to EEG-visual alignment, advancing both performance and interpretability. Future work may incorporate eye-tracking-based validation and extend the framework to multi-subject modeling, single-trial decoding, and real-time BCI scenarios.

\begin{ack}
This work was supported in part by Natural Science Foundation of China under Grant 62472220, Grant 62371234 and Grant 62076129.
\end{ack}

\bibliography{example_paper}

\newpage
\appendix

\section{Supplementary Material}
\subsection{Details of the Datasets}
\label{app:datasets}
\textbf{THINGS-EEG} \cite{t14} is a large-scale neuroimaging resource designed to facilitate zero-shot visual object recognition from neural signals. The dataset comprises electroencephalography recordings from ten healthy adult participants with normal or corrected-to-normal vision. EEG signals were acquired using a 64-channel EASYCAP arranged according to the standard 10–10 system and a BrainVision actiCHamp amplifier, sampled at 1,000 Hz with online filtering between 0.1 and 100 Hz and referenced to the Fz electrode. Visual stimulation followed a Rapid Serial Visual Presentation (RSVP) protocol. An orthogonal detection task was employed to ensure sustained attention. After excluding target trials and retaining the specified repetitions, the training and test sets contain 82,160 trials per subject in total. The stimulus set was drawn from the THINGS image database, encompassing 1,854 distinct object concepts organized into 27 higher-level categories (e.g., animals, vehicles, food items, tools, sports equipment). These concepts were partitioned into 1,654 training categories (10 images per category, 4 repetitions each) and 200 hold-out test categories (1 image per category, 80 repetitions for signal averaging).

\textbf{THINGS-MEG} \cite{t52} contains magnetoencephalography recordings from four healthy participants, with 271 MEG channels retained for analysis. The experimental protocol involved 500 ms stimulus presentation followed by a 1,000 ± 200 ms blank interval, with participants performing an oddball detection task. The dataset comprises 1,654 training concepts (12 images per category, single presentation) and 200 test concepts (1 image per category, 12 repetitions). MEG data were band-pass filtered (0.1–100 Hz), epoched from 0–250 ms post-stimulus, and downsampled to 200 Hz, with trial averaging applied to enhance sensor-level signal stability.

\subsection{Comparison Methods}
BraVL \cite{t6} develops a multimodal variational autoencoding framework based on mixture-of-product-of-experts to jointly embed brain activity, visual features, and textual descriptions into a shared latent representation. It maximizes both intra- and inter-modality mutual information to ensure consistent joint generation and improve data efficiency in zero-shot neural decoding scenarios.

NICE \cite{t16} presents a self-supervised framework that aligns image and EEG representations via contrastive learning for zero-shot object recognition. It maximizes the cosine similarity between paired image stimuli and EEG responses while pushing non-paired samples apart, enabling the EEG encoder to extract semantics consistent with visual features without categorical labels. Additionally, plug-and-play self-attention and graph attention modules are introduced to capture spatial correlations among EEG channels, supporting the analysis of spatial contributions to visual decoding.

ATM \cite{t17} proposes an end-to-end zero-shot visual reconstruction framework that projects EEG signals into a shared embedding space aligned with CLIP. It employs channel-wise Transformer encoders with temporal-spatial convolution to capture semantic and spatial correlations, facilitating tasks like classification, retrieval, and image generation.

CogCap \cite{t43} presents a unified multimodal decoding framework that incorporates text and depth alongside EEG and image modalities. It uses dedicated encoders to align EEG with each modality, capturing “beyond-image” information, and a lightweight diffusion prior to map EEG embeddings into the CLIP space for high-fidelity reconstruction.

VE-SDN \cite{t47} disentangles semantic and domain-specific features through mutual-information-based semantic decoupling and cross-modal alignment. It constructs a shared semantic space for visual and EEG representations and introduces an intra-class geometric consistency constraint to improve semantic consistency.

MB2C \cite{t51} employs a dual-GAN architecture with cycle consistency loss and contrastive learning to generate and inversely translate modality-related features. This approach narrows the modality gap between EEG and visual features, ensuring semantically aligned embeddings reside in a shared representation space.

Neural-MCRL \cite{t50} achieves multimodal alignment via semantic bridging and cross-attention mechanisms, ensuring intra-modal semantic completion and inter-modal consistency. It introduces NESTA to adaptively capture spectral-temporal EEG patterns with subject-specific transformations for robust visual decoding.

UBP \cite{t20} proposes an Uncertainty-aware Blur Prior framework that quantifies uncertainty within paired brain-visual data to estimate the inherent mismatch. It dynamically blurs high-frequency details of visual stimuli accordingly, thereby reducing information discrepancy and enhancing alignment robustness for zero-shot neural decoding.

\subsection{Statistical Robustness and Capacity Control}
\label{app:robustness_capacity}

\subsubsection{Multi-Seed Evaluation}

We evaluate CAIA under five random seeds in the subject-dependent, 200-way zero-shot retrieval setting on THINGS-EEG (\cref{tab:multi_seed_results}).

\begin{table}[htbp]
  \centering
  \caption{CAIA Top-1/Top-5 accuracy (\%) across five random seeds on THINGS-EEG.}
  \label{tab:multi_seed_results}
  \resizebox{\linewidth}{!}{%
  \begin{tabular}{llccccccccccc}
    \toprule
    Seed & Metric & Sub 1 & Sub 2 & Sub 3 & Sub 4 & Sub 5 & Sub 6 & Sub 7 & Sub 8 & Sub 9 & Sub 10 & Avg \\
    \midrule
    0 & Top-1 & 74.0 & 80.0 & 69.0 & 51.0 & 67.5 & 67.5 & 63.0 & 83.0 & 69.0 & 79.0 & 70.30 \\
      & Top-5 & 98.0 & 97.5 & 94.0 & 92.5 & 94.0 & 96.5 & 95.5 & 99.0 & 96.5 & 97.5 & 96.10 \\
    \midrule
    1 & Top-1 & 72.5 & 78.0 & 66.0 & 48.5 & 64.5 & 65.0 & 61.0 & 80.0 & 67.5 & 76.5 & 67.95 \\
      & Top-5 & 97.5 & 97.0 & 93.5 & 91.5 & 93.0 & 95.5 & 94.5 & 98.0 & 96.0 & 96.5 & 95.30 \\
    \midrule
    2 & Top-1 & 75.5 & 83.0 & 70.5 & 54.5 & 68.0 & 66.5 & 66.0 & 85.5 & 71.0 & 82.0 & 72.25 \\
      & Top-5 & 98.0 & 98.0 & 94.5 & 93.0 & 94.5 & 96.0 & 96.0 & 99.0 & 97.0 & 98.0 & 96.40 \\
    \midrule
    3 & Top-1 & 70.0 & 77.0 & 65.0 & 47.0 & 62.5 & 64.5 & 60.0 & 80.0 & 65.5 & 75.0 & 66.65 \\
      & Top-5 & 97.0 & 96.5 & 93.0 & 91.0 & 92.5 & 95.0 & 94.0 & 98.0 & 95.5 & 96.0 & 94.85 \\
    \midrule
    4 & Top-1 & 73.0 & 81.5 & 68.0 & 52.0 & 70.0 & 65.5 & 64.5 & 84.0 & 72.0 & 80.5 & 71.10 \\
      & Top-5 & 97.5 & 98.0 & 94.0 & 92.5 & 95.0 & 96.0 & 95.5 & 98.5 & 97.0 & 97.5 & 96.15 \\
    \bottomrule
  \end{tabular}%
  }
\end{table}

Across five seeds, CAIA achieves Top-1 and Top-5 accuracies of 69.65\% $\pm$ 1.03\% and 95.76\% $\pm$ 0.29\% (mean $\pm$ SE), with 95\% confidence intervals of [66.79\%, 72.51\%] and [94.95\%, 96.57\%], respectively. These results support the stability of the reported retrieval performance across random initializations.

\subsubsection{Paired Comparison with UBP}

We additionally compare CAIA with UBP using seed-matched runs at seed 4. The ten subjects serve as paired observations, and we apply two-sided paired $t$-tests to their accuracy differences. As reported in \cref{tab:seed_matched_results}, CAIA improves mean Top-1 and Top-5 accuracy by 21.40 and 17.60 percentage points, respectively. The differences are statistically significant for both Top-1 ($t(9)=7.13$, $p<0.0001$) and Top-5 ($t(9)=11.35$, $p<0.0001$), supporting the improvement over UBP across subjects under the matched seed.

\begin{table}[htbp]
  \centering
  \caption{Seed-matched Top-1/Top-5 accuracy (\%) of UBP and CAIA on THINGS-EEG at seed 4.}
  \label{tab:seed_matched_results}
  \resizebox{\linewidth}{!}{%
  \begin{tabular}{llccccccccccc}
    \toprule
    Method & Metric & Sub 1 & Sub 2 & Sub 3 & Sub 4 & Sub 5 & Sub 6 & Sub 7 & Sub 8 & Sub 9 & Sub 10 & Avg \\
    \midrule
    UBP & Top-1 & 42.0 & 49.5 & 49.5 & 48.5 & 42.5 & 56.0 & 48.0 & 56.5 & 44.5 & 60.0 & 49.70 \\
        & Top-5 & 70.5 & 79.5 & 81.0 & 75.5 & 72.5 & 82.0 & 78.5 & 84.0 & 75.5 & 86.5 & 78.55 \\
    \midrule
    CAIA & Top-1 & 73.0 & 81.5 & 68.0 & 52.0 & 70.0 & 65.5 & 64.5 & 84.0 & 72.0 & 80.5 & 71.10 \\
         & Top-5 & 97.5 & 98.0 & 94.0 & 92.5 & 95.0 & 96.0 & 95.5 & 98.5 & 97.0 & 97.5 & 96.15 \\
    \bottomrule
  \end{tabular}%
  }
\end{table}

\subsubsection{Capacity-Matched Comparison}

To examine whether increased model capacity accounts for the performance gain, we expand the EEG projection head of UBP to approximately 68 million trainable parameters without adding the proposed CAIA components. The resulting variant achieves 51.35\% Top-1 and 80.20\% Top-5 accuracy (\cref{tab:capacity_matched_results}), remaining substantially below CAIA's 70.30\% and 96.10\%. This comparison suggests that increasing encoder capacity alone does not account for the observed improvement.

\begin{table}[htbp]
  \centering
  \caption{Capacity-matched UBP Top-1/Top-5 accuracy (\%) on THINGS-EEG.}
  \label{tab:capacity_matched_results}
  \resizebox{\linewidth}{!}{%
  \begin{tabular}{lccccccccccc}
    \toprule
    Metric & Sub 1 & Sub 2 & Sub 3 & Sub 4 & Sub 5 & Sub 6 & Sub 7 & Sub 8 & Sub 9 & Sub 10 & Avg \\
    \midrule
    Top-1 & 44.0 & 51.5 & 51.5 & 49.5 & 44.5 & 58.0 & 49.0 & 58.5 & 46.0 & 61.0 & 51.35 \\
    Top-5 & 72.5 & 81.5 & 83.0 & 76.5 & 74.5 & 83.5 & 80.0 & 85.5 & 77.5 & 87.5 & 80.20 \\
    \bottomrule
  \end{tabular}%
  }
\end{table}

\subsection{Detailed Ablation Results}
\label{app:detailed_ablation}

\cref{tab:full_ablation} reports the per-subject results for all eight combinations of $IBWave$, $ATTN$, and $L_{\text{bound}}$, supplementing the ablation analysis in the main text.

\begin{table}[htbp]
  \centering
  \caption{Ablation study results (\%) of 200-way zero-shot retrieval on THINGS-EEG: Top-1 and Top-5.}
  \label{tab:full_ablation}
  \resizebox{\linewidth}{!}{%
  \begin{tabular}{ccccccccccccccccccccccccc}
    \toprule
    \multicolumn{3}{c}{Modules}
    & \multicolumn{2}{c}{Sub 1}
    & \multicolumn{2}{c}{Sub 2}
    & \multicolumn{2}{c}{Sub 3}
    & \multicolumn{2}{c}{Sub 4}
    & \multicolumn{2}{c}{Sub 5}
    & \multicolumn{2}{c}{Sub 6}
    & \multicolumn{2}{c}{Sub 7}
    & \multicolumn{2}{c}{Sub 8}
    & \multicolumn{2}{c}{Sub 9}
    & \multicolumn{2}{c}{Sub 10}
    & \multicolumn{2}{c}{Avg} \\
    \cmidrule(r){1-3}
    \cmidrule(r){4-5}
    \cmidrule(r){6-7}
    \cmidrule(r){8-9}
    \cmidrule(r){10-11}
    \cmidrule(r){12-13}
    \cmidrule(r){14-15}
    \cmidrule(r){16-17}
    \cmidrule(r){18-19}
    \cmidrule(r){20-21}
    \cmidrule(r){22-23}
    \cmidrule(r){24-25}
    IBWave & ATTN & $L_{\text{bound}}$
    & T1 & T5 & T1 & T5 & T1 & T5 & T1 & T5 & T1 & T5
    & T1 & T5 & T1 & T5 & T1 & T5 & T1 & T5 & T1 & T5
    & T1 & T5 \\
    \midrule
    $\times$ & $\times$ & $\times$ & 51.5 & 82.0 & 55.0 & 89.0 & 58.0 & 89.5 & 45.0 & 82.0 & 45.5 & 79.0 & 57.5 & 92.0 & 59.0 & 91.5 & 64.5 & 92.5 & 47.5 & 83.0 & 61.5 & 92.5 & 54.5 & 87.3 \\
    $\times$ & $\times$ & \checkmark & 51.5 & 85.0 & 43.5 & 80.5 & 49.5 & 84.5 & 48.0 & 83.0 & 42.0 & 81.0 & 56.5 & 90.5 & 59.5 & 87.5 & 62.5 & 93.0 & 42.0 & 80.0 & 68.5 & 91.5 & 52.4 & 85.7 \\
    $\times$ & \checkmark & $\times$ & 60.5 & 91.0 & 66.0 & 94.5 & 61.5 & 93.5 & \underline{54.0} & \underline{90.5} & 52.0 & 84.0 & 68.5 & \underline{95.5} & 60.5 & \underline{93.0} & 67.5 & 95.5 & 58.5 & 91.5 & 71.5 & 96.5 & 62.1 & 92.6 \\
    $\times$ & \checkmark & \checkmark & 66.5 & 92.5 & \underline{70.0} & 94.5 & \underline{66.5} & \textbf{95.0} & \textbf{54.5} & 89.5 & 52.0 & 87.0 & \underline{69.5} & 95.5 & 61.5 & 91.0 & 69.5 & 95.5 & 59.0 & 92.0 & 70.5 & \underline{97.0} & 64.0 & 93.0 \\
    \checkmark & $\times$ & $\times$ & 66.5 & 93.5 & 63.0 & 93.5 & 61.0 & 90.5 & 49.0 & 84.5 & 59.5 & 86.5 & 68.0 & 95.0 & \textbf{65.5} & 92.5 & \underline{75.5} & 98.0 & \underline{61.0} & 88.5 & 70.5 & 93.5 & 64.0 & 91.6 \\
    \checkmark & $\times$ & \checkmark & \underline{68.0} & \underline{95.5} & 69.0 & 93.5 & 62.0 & 90.5 & 51.0 & 85.0 & \underline{62.0} & 89.0 & 69.5 & 95.5 & 61.5 & 91.5 & 74.5 & 97.5 & 60.0 & 90.5 & 71.0 & 95.0 & \underline{64.9} & 92.4 \\
    \checkmark & \checkmark & $\times$ & 66.5 & 94.5 & 68.5 & \underline{95.0} & 66.5 & 94.0 & 53.5 & 89.0 & 57.0 & \underline{90.0} & \textbf{70.0} & 95.5 & 61.0 & 92.0 & 73.5 & \underline{98.5} & 60.0 & \underline{92.5} & \underline{72.5} & 95.0 & 64.9 & \underline{93.6} \\
    \rowcolor{blue!10}
    \checkmark & \checkmark & \checkmark & \textbf{74.0} & \textbf{98.0} & \textbf{80.0} & \textbf{97.5} & \textbf{69.0} & \underline{94.0} & 51.0 & \textbf{92.5} & \textbf{67.5} & \textbf{94.0} & 67.5 & \textbf{96.5} & \underline{63.0} & \textbf{95.5} & \textbf{83.0} & \textbf{99.0} & \textbf{69.0} & \textbf{96.5} & \textbf{79.0} & \textbf{97.5} & \textbf{70.3} & \textbf{96.1} \\
    \bottomrule
  \end{tabular}%
  }
\end{table}

\subsection{Adaptive Visual Fusion Analysis}
\label{app:visual_fusion}

\subsubsection{Image-Fusion Strategy Comparison}

\cref{tab:full_image_fusion} reports the per-subject comparison of image-fusion strategies. EEG-guided attention achieves the highest mean Top-1 and Top-5 accuracy, outperforming individual blurring pathways, fixed linear blending, and a global learnable mask. These results support adapting the balance between center-biased and saliency-guided processing to the neural context.

\begin{table}[htbp]
  \centering
  \caption{Image fusion method comparison results (\%) of 200-way zero-shot retrieval on THINGS-EEG: Top-1 and Top-5 (intra-subject).}
  \label{tab:full_image_fusion}
  \resizebox{\linewidth}{!}{%
  \begin{tabular}{lcccccccccccccccccccccc}
    \toprule
    & \multicolumn{2}{c}{Sub 1} & \multicolumn{2}{c}{Sub 2} & \multicolumn{2}{c}{Sub 3} & \multicolumn{2}{c}{Sub 4} & \multicolumn{2}{c}{Sub 5} & \multicolumn{2}{c}{Sub 6} & \multicolumn{2}{c}{Sub 7} & \multicolumn{2}{c}{Sub 8} & \multicolumn{2}{c}{Sub 9} & \multicolumn{2}{c}{Sub 10} & \multicolumn{2}{c}{Avg} \\
    \cmidrule(r){2-3} \cmidrule(r){4-5} \cmidrule(r){6-7} \cmidrule(r){8-9} \cmidrule(r){10-11} \cmidrule(r){12-13} \cmidrule(r){14-15} \cmidrule(r){16-17} \cmidrule(r){18-19} \cmidrule(r){20-21} \cmidrule(r){22-23}
    Method & T1 & T5 & T1 & T5 & T1 & T5 & T1 & T5 & T1 & T5 & T1 & T5 & T1 & T5 & T1 & T5 & T1 & T5 & T1 & T5 & T1 & T5 \\
    \midrule
    Original & 67.0 & 91.5 & 64.0 & \underline{94.0} & \textbf{70.0} & 94.0 & 45.0 & 84.0 & \underline{66.5} & 88.0 & 72.0 & 95.0 & 59.5 & 90.5 & 73.0 & 95.0 & \underline{63.0} & \underline{93.0} & 66.0 & \underline{96.0} & 64.6 & 92.1 \\
    Center\_blur & \underline{68.0} & \underline{95.5} & \underline{69.0} & 93.5 & 62.0 & 90.5 & \textbf{51.0} & \underline{85.0} & 62.0 & \underline{89.0} & 69.5 & \underline{95.5} & 61.5 & 91.5 & \underline{74.5} & \underline{97.5} & 60.0 & 90.5 & \underline{71.0} & 95.0 & \underline{64.9} & \underline{92.4} \\
    Saliency\_blur & 60.0 & 88.5 & 54.5 & 88.0 & 62.5 & 92.5 & 47.5 & 80.5 & 51.0 & 83.5 & 70.0 & 90.0 & 60.5 & 90.0 & 67.5 & 95.0 & 50.5 & 80.0 & 68.5 & 92.0 & 59.3 & 88.0 \\
    Linear & 66.0 & 91.0 & 62.0 & 93.5 & 66.0 & 93.5 & 47.5 & 81.5 & 60.5 & 84.5 & \underline{72.5} & 95.5 & \textbf{65.5} & \textbf{95.5} & 68.5 & 96.5 & 60.0 & 91.0 & 63.0 & 92.5 & 63.2 & 91.5 \\
    Learnable\_mask & 66.0 & 93.0 & 62.0 & 94.0 & \textbf{70.0} & \textbf{95.0} & 46.0 & 82.0 & 61.5 & 87.0 & \textbf{74.5} & 95.0 & \underline{65.0} & 95.0 & 68.0 & 94.0 & 61.5 & 92.0 & 66.5 & 93.5 & 64.1 & 92.1 \\
    \rowcolor{blue!10}
    \textbf{Attention (Ours)} & \textbf{74.0} & \textbf{98.0} & \textbf{80.0} & \textbf{97.5} & 69.0 & \underline{94.0} & \textbf{51.0} & \textbf{92.5} & \textbf{67.5} & \textbf{94.0} & 67.5 & \textbf{96.5} & 63.0 & \textbf{95.5} & \textbf{83.0} & \textbf{99.0} & \textbf{69.0} & \textbf{96.5} & \textbf{79.0} & \textbf{97.5} & \textbf{70.3} & \textbf{96.1} \\
    \bottomrule
  \end{tabular}%
  }
\end{table}

\subsubsection{Saliency Detection Model Comparison}

We evaluate three saliency detection backends in the saliency-guided blurring branch (\cref{tab:saliency_comparison}). U\textsuperscript{2}Net is a deep salient-object detector pretrained on generic saliency benchmarks. Spectral Residual and Fine-Grained are provided by OpenCV and derive saliency from spectral irregularities and local contrast, respectively.

Fine-Grained achieves the highest mean accuracy (70.3\% Top-1, 96.1\% Top-5). Its contrast-based saliency may provide a useful balance between local visual detail and broader image structure, guiding adaptive blurring to preserve distinctive features while reducing background detail. The lower performance of U\textsuperscript{2}Net may reflect a mismatch between its generic salient-object predictions and the visual detail useful for EEG-image alignment. Spectral Residual may place greater emphasis on isolated spectral irregularities, yielding preservation priorities that are less suitable for this decoding task.

\begin{table}[htbp]
  \centering
  \caption{Comparison of different saliency detection models within CAIA on THINGS-EEG (200-way zero-shot retrieval, subject-dependent, Top-1/Top-5, \%).}
  \label{tab:saliency_comparison}
  \resizebox{\linewidth}{!}{%
  \begin{tabular}{lcccccccccccccccccccccc}
    \toprule
    & \multicolumn{2}{c}{Sub 1} & \multicolumn{2}{c}{Sub 2} & \multicolumn{2}{c}{Sub 3} & \multicolumn{2}{c}{Sub 4} & \multicolumn{2}{c}{Sub 5} & \multicolumn{2}{c}{Sub 6} & \multicolumn{2}{c}{Sub 7} & \multicolumn{2}{c}{Sub 8} & \multicolumn{2}{c}{Sub 9} & \multicolumn{2}{c}{Sub 10} & \multicolumn{2}{c}{Avg} \\
    \cmidrule(r){2-3} \cmidrule(r){4-5} \cmidrule(r){6-7} \cmidrule(r){8-9} \cmidrule(r){10-11} \cmidrule(r){12-13} \cmidrule(r){14-15} \cmidrule(r){16-17} \cmidrule(r){18-19} \cmidrule(r){20-21} \cmidrule(r){22-23}
    Method & T1 & T5 & T1 & T5 & T1 & T5 & T1 & T5 & T1 & T5 & T1 & T5 & T1 & T5 & T1 & T5 & T1 & T5 & T1 & T5 & T1 & T5 \\
    \midrule
    U\textsuperscript{2}Net & \underline{67.5} & \underline{95.5} & \underline{75.0} & \underline{96.5} & \textbf{73.0} & \textbf{96.5} & \textbf{55.5} & \underline{91.0} & 63.0 & 92.0 & \textbf{75.0} & \textbf{96.5} & \underline{62.0} & \underline{94.5} & \underline{80.5} & \underline{97.5} & \underline{63.0} & 93.0 & \underline{77.5} & \underline{97.0} & \underline{69.2} & \underline{95.0} \\
    Spectral Residual & 62.0 & 91.0 & 71.5 & 96.0 & \underline{71.0} & \underline{96.0} & 47.5 & 88.0 & \underline{66.5} & \textbf{95.0} & \underline{71.0} & 96.5 & 62.0 & 94.0 & 78.0 & 97.5 & 62.0 & \underline{94.0} & 74.5 & 96.5 & 66.6 & 94.5 \\
    \rowcolor{blue!10}
    Fine-Grained & \textbf{74.0} & \textbf{98.0} & \textbf{80.0} & \textbf{97.5} & 69.0 & 94.0 & \underline{51.0} & \textbf{92.5} & \textbf{67.5} & \underline{94.0} & 67.5 & \textbf{96.5} & \textbf{63.0} & \textbf{95.5} & \textbf{83.0} & \textbf{99.0} & \textbf{69.0} & \textbf{96.5} & \textbf{79.0} & \textbf{97.5} & \textbf{70.3} & \textbf{96.1} \\
    \bottomrule
  \end{tabular}%
  }
\end{table}

\subsubsection{EEG-Shuffling Analysis}

To examine the role of EEG conditioning in adaptive fusion, we replace the EEG input to the cross-modal attention module with that of a randomly selected different sample within the batch, while all other components retain their original inputs. This perturbation reduces mean Top-1 and Top-5 accuracy from 70.3\% / 96.1\% to 64.5\% / 92.0\% (\cref{tab:shuffled_eeg_results}). The spatial correlation between attention maps obtained with paired and shuffled EEG is 0.0057.

\begin{figure}[htbp]
  \centering
  \includegraphics[width=0.6\linewidth]{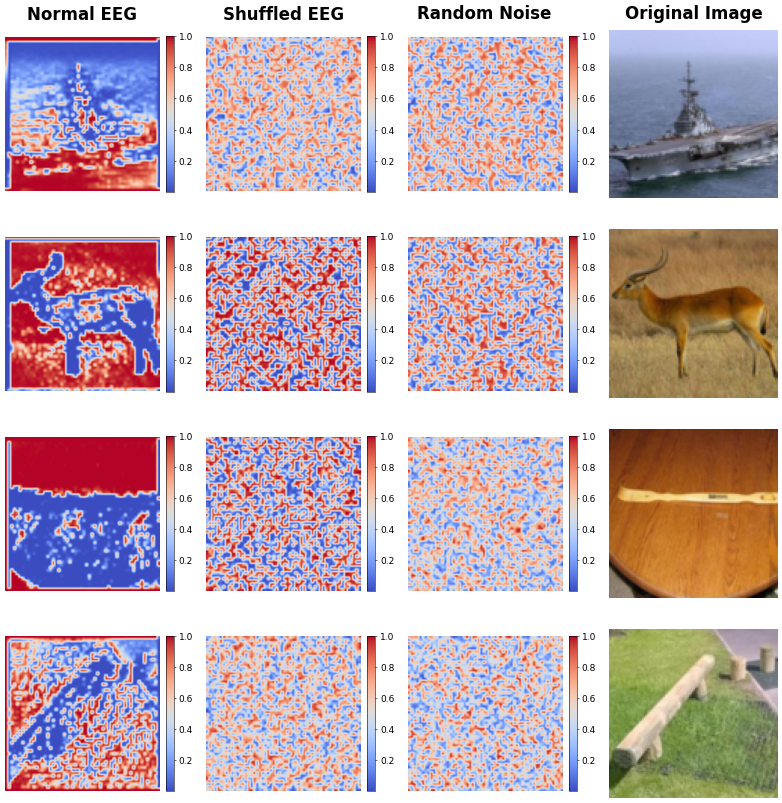}
  \caption{Attention maps with paired EEG, shuffled EEG, and random noise, alongside the corresponding stimulus images.}
  \label{fig:attention_shuffle_comparison}
\end{figure}

As shown in \cref{fig:attention_shuffle_comparison}, paired EEG produces spatially structured fusion patterns, whereas shuffled EEG and random noise yield more fragmented maps. These observations support the contribution of sample-specific neural context to adaptive visual fusion.

\begin{table}[htbp]
  \centering
  \caption{Top-1/Top-5 accuracy (\%) with shuffled EEG inputs to the attention module on THINGS-EEG.}
  \label{tab:shuffled_eeg_results}
  \resizebox{\linewidth}{!}{%
  \begin{tabular}{lccccccccccc}
    \toprule
    Metric & Sub 1 & Sub 2 & Sub 3 & Sub 4 & Sub 5 & Sub 6 & Sub 7 & Sub 8 & Sub 9 & Sub 10 & Avg \\
    \midrule
    Top-1 & 66.5 & 68.0 & 63.5 & 50.0 & 63.0 & 68.0 & 60.0 & 75.0 & 60.5 & 70.0 & 64.5 \\
    Top-5 & 94.5 & 94.0 & 91.0 & 84.0 & 88.0 & 94.5 & 92.0 & 97.0 & 91.0 & 94.0 & 92.0 \\
    \bottomrule
  \end{tabular}%
  }
\end{table}

\subsubsection{Blur Strength Sensitivity}

Moderate blurring ($w_0=0.5$) achieves the highest mean accuracy, while lighter ($w_0=0.3$) and heavier ($w_0=0.7$) blurring yield lower performance (\cref{tab:w0_sensitivity}). This pattern supports balancing visual information reduction with semantic preservation.

\begin{table}[htbp]
  \centering
  \caption{Sensitivity analysis of the blur strength parameter $w_0$ on THINGS-EEG (200-way zero-shot retrieval, subject-dependent, Top-1/Top-5, \%).}
  \label{tab:w0_sensitivity}
  \resizebox{\linewidth}{!}{%
  \begin{tabular}{lcccccccccccccccccccccc}
    \toprule
    & \multicolumn{2}{c}{Sub 1} & \multicolumn{2}{c}{Sub 2} & \multicolumn{2}{c}{Sub 3} & \multicolumn{2}{c}{Sub 4} & \multicolumn{2}{c}{Sub 5} & \multicolumn{2}{c}{Sub 6} & \multicolumn{2}{c}{Sub 7} & \multicolumn{2}{c}{Sub 8} & \multicolumn{2}{c}{Sub 9} & \multicolumn{2}{c}{Sub 10} & \multicolumn{2}{c}{Avg} \\
    \cmidrule(r){2-3} \cmidrule(r){4-5} \cmidrule(r){6-7} \cmidrule(r){8-9} \cmidrule(r){10-11} \cmidrule(r){12-13} \cmidrule(r){14-15} \cmidrule(r){16-17} \cmidrule(r){18-19} \cmidrule(r){20-21} \cmidrule(r){22-23}
    $w_0$ & T1 & T5 & T1 & T5 & T1 & T5 & T1 & T5 & T1 & T5 & T1 & T5 & T1 & T5 & T1 & T5 & T1 & T5 & T1 & T5 & T1 & T5 \\
    \midrule
    Light ($w_0=0.3$) & \underline{68.0} & 93.0 & \underline{77.0} & 94.0 & \textbf{82.0} & \textbf{99.0} & 50.0 & \textbf{97.0} & \underline{60.5} & \underline{92.5} & \textbf{80.5} & \textbf{97.5} & \underline{58.0} & \underline{93.0} & \textbf{85.0} & \underline{98.5} & 57.5 & \underline{94.5} & 76.0 & \textbf{97.5} & \underline{69.5} & \underline{95.7} \\
    Heavy ($w_0=0.7$) & 68.0 & \underline{95.0} & 77.0 & \underline{97.0} & 69.0 & 92.5 & \textbf{58.0} & \underline{95.0} & 59.0 & 92.0 & \underline{70.5} & 95.5 & 57.0 & 92.0 & 74.5 & 95.5 & \underline{58.5} & 92.0 & \underline{77.0} & 94.5 & 66.9 & 94.1 \\
    \rowcolor{blue!10}
    \textbf{Standard ($w_0=0.5$)} & \textbf{74.0} & \textbf{98.0} & \textbf{80.0} & \textbf{97.5} & \underline{69.0} & \underline{94.0} & \underline{51.0} & 92.5 & \textbf{67.5} & \textbf{94.0} & 67.5 & \underline{96.5} & \textbf{63.0} & \textbf{95.5} & \underline{83.0} & \textbf{99.0} & \textbf{69.0} & \textbf{96.5} & \textbf{79.0} & \textbf{97.5} & \textbf{70.3} & \textbf{96.1} \\
    \bottomrule
  \end{tabular}%
  }
\end{table}

\subsection{Frequency Screening Analysis}
\label{app:frequency_screening}

\subsubsection{Frequency Filter Comparison}

We compare IBWave against three static band-pass filters (\cref{tab:freq_filter_comparison}): a high-frequency filter ($>40$ Hz), a low-frequency filter ($<8$ Hz), and a mid-frequency filter (8--40 Hz).

Among the static filters, the mid-frequency filter achieves the highest mean accuracy (58.1\% Top-1, 90.1\% Top-5), consistent with the Beta-dominant weighting learned by our model and its reported involvement in visual attention. However, its accuracy remains substantially below that of IBWave (70.3\% Top-1, 96.1\% Top-5). Static filters rely on fixed frequency boundaries, whereas IBWave jointly optimizes decomposition scales and band-selection weights through the decoding objective. This allows the spectral emphasis to adapt to the information useful for neural-visual alignment. These results support the benefit of learnable, task-adaptive frequency screening over fixed band-pass filtering.

\begin{table}[htbp]
  \centering
  \caption{Comparison of IBWave with static band-pass filters on THINGS-EEG (200-way zero-shot retrieval, subject-dependent, Top-1/Top-5, \%).}
  \label{tab:freq_filter_comparison}
  \resizebox{\linewidth}{!}{%
  \begin{tabular}{lcccccccccccccccccccccc}
    \toprule
    & \multicolumn{2}{c}{Sub 1} & \multicolumn{2}{c}{Sub 2} & \multicolumn{2}{c}{Sub 3} & \multicolumn{2}{c}{Sub 4} & \multicolumn{2}{c}{Sub 5} & \multicolumn{2}{c}{Sub 6} & \multicolumn{2}{c}{Sub 7} & \multicolumn{2}{c}{Sub 8} & \multicolumn{2}{c}{Sub 9} & \multicolumn{2}{c}{Sub 10} & \multicolumn{2}{c}{Avg} \\
    \cmidrule(r){2-3} \cmidrule(r){4-5} \cmidrule(r){6-7} \cmidrule(r){8-9} \cmidrule(r){10-11} \cmidrule(r){12-13} \cmidrule(r){14-15} \cmidrule(r){16-17} \cmidrule(r){18-19} \cmidrule(r){20-21} \cmidrule(r){22-23}
    Method & T1 & T5 & T1 & T5 & T1 & T5 & T1 & T5 & T1 & T5 & T1 & T5 & T1 & T5 & T1 & T5 & T1 & T5 & T1 & T5 & T1 & T5 \\
    \midrule
    High-Frequency & 22.0 & 46.5 & 24.5 & 63.0 & 11.5 & 43.0 & 18.5 & 48.5 & 9.0 & 29.5 & 20.5 & 53.0 & 23.0 & 57.0 & 24.5 & 58.5 & 23.0 & 56.5 & 27.5 & 58.0 & 20.4 & 51.4 \\
    Low-Frequency & 43.5 & 86.5 & 38.5 & 78.5 & 48.0 & 88.5 & 39.0 & 82.5 & 44.5 & 84.0 & 44.5 & 87.5 & \underline{52.0} & \underline{87.0} & \underline{61.5} & \underline{91.5} & 40.0 & \underline{84.0} & 53.5 & \underline{95.5} & 46.5 & 86.6 \\
    Mid-Frequency & \underline{62.0} & \underline{93.5} & \underline{63.5} & \underline{93.0} & \underline{65.5} & \textbf{94.0} & \textbf{52.5} & \underline{86.5} & \underline{59.0} & \underline{86.5} & \underline{66.0} & \underline{94.5} & 44.5 & 84.0 & 51.5 & 89.5 & \underline{45.5} & 84.0 & \underline{70.5} & 95.0 & \underline{58.1} & \underline{90.1} \\
    \rowcolor{blue!10}
    \textbf{IBWave (Ours)} & \textbf{74.0} & \textbf{98.0} & \textbf{80.0} & \textbf{97.5} & \textbf{69.0} & \textbf{94.0} & \underline{51.0} & \textbf{92.5} & \textbf{67.5} & \textbf{94.0} & \textbf{67.5} & \textbf{96.5} & \textbf{63.0} & \textbf{95.5} & \textbf{83.0} & \textbf{99.0} & \textbf{69.0} & \textbf{96.5} & \textbf{79.0} & \textbf{97.5} & \textbf{70.3} & \textbf{96.1} \\
    \bottomrule
  \end{tabular}%
  }
\end{table}

\subsubsection{Entropy Regularization Ablation}

To examine the contribution of entropy regularization, we retain the learnable decomposition scales and band-selection weights while removing the entropy term. This variant achieves 67.8\% Top-1 and 94.2\% Top-5 accuracy, compared with 70.3\% and 96.1\% for the full model (\cref{tab:entropy_ablation}). The decrease supports the role of entropy regularization in encouraging selective spectral weighting for visual decoding.

\begin{table}[htbp]
  \centering
  \caption{Top-1/Top-5 accuracy (\%) without entropy regularization on THINGS-EEG.}
  \label{tab:entropy_ablation}
  \resizebox{\linewidth}{!}{%
  \begin{tabular}{lccccccccccc}
    \toprule
    Metric & Sub 1 & Sub 2 & Sub 3 & Sub 4 & Sub 5 & Sub 6 & Sub 7 & Sub 8 & Sub 9 & Sub 10 & Avg \\
    \midrule
    Top-1 & 71.0 & 75.0 & 68.5 & 56.5 & 60.5 & 68.0 & 63.0 & 74.5 & 65.5 & 75.0 & 67.8 \\
    Top-5 & 95.0 & 96.0 & 94.0 & 91.5 & 90.5 & 95.0 & 92.5 & 96.5 & 94.0 & 97.0 & 94.2 \\
    \bottomrule
  \end{tabular}%
  }
\end{table}

\subsubsection{Spectral Preference and Weight Stability}

\begin{figure}[htbp]
  \centering
  \begin{minipage}[t]{0.48\textwidth}
    \centering
    \includegraphics[width=\linewidth]{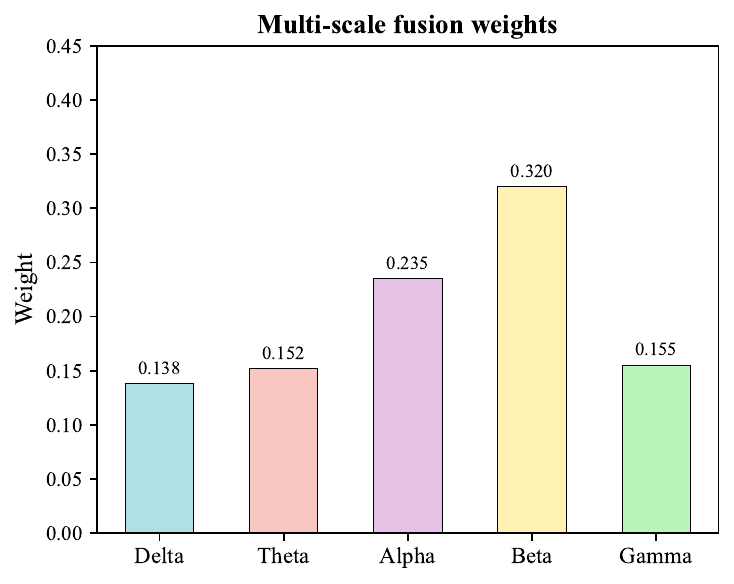}
    \caption{Band-selection weights in the subject-dependent setting.}
    \label{scale}
  \end{minipage}
  \hfill
  \begin{minipage}[t]{0.48\textwidth}
    \centering
    \includegraphics[width=\linewidth]{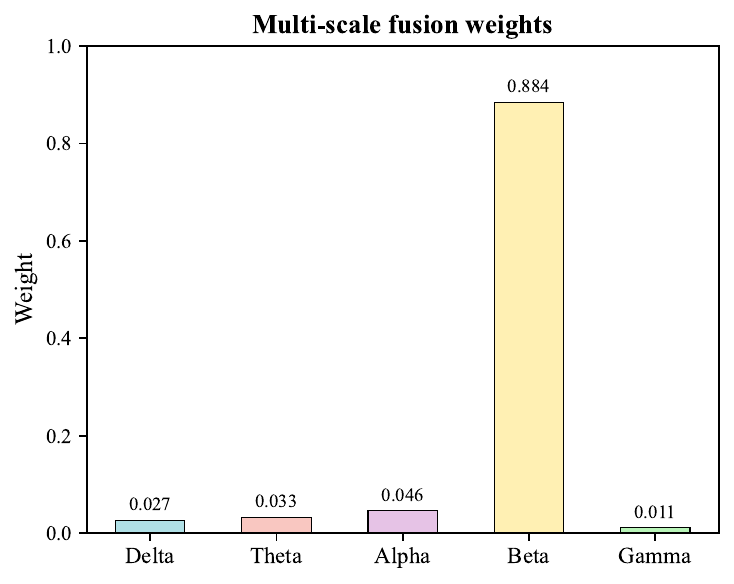}
    \caption{Band-selection weights in the subject-independent setting.}
    \label{inter_scale}
  \end{minipage}
\end{figure}

The learned band-selection weights exhibit a preference for the Beta band (\cref{scale,inter_scale}), consistent with its reported involvement in visual processing and attention \cite{t53,t54}. Across seeds, within-subject weight variation remains no greater than 0.005 for all bands, while between-subject differences are larger, reaching 0.038 for the Beta band. This pattern suggests a reproducible spectral preference alongside subject-specific adaptation. The observed weight ranges are reported in \cref{tab:band_weight_ranges}.

\begin{table}[htbp]
  \centering
  \caption{Ranges of learned band-selection weights across subjects and seeds.}
  \label{tab:band_weight_ranges}
  \begin{tabular}{lc}
    \toprule
    Band & Weight Range \\
    \midrule
    Delta ($\delta$) & 0.126--0.152 \\
    Theta ($\theta$) & 0.133--0.161 \\
    Alpha ($\alpha$) & 0.218--0.248 \\
    Beta ($\beta$)   & 0.302--0.340 \\
    Gamma ($\gamma$) & 0.144--0.163 \\
    \bottomrule
  \end{tabular}
\end{table}

\subsubsection{Subject-Independent Screening Analysis}

\begin{figure}[htbp]
  \centering
  \includegraphics[width=\linewidth]{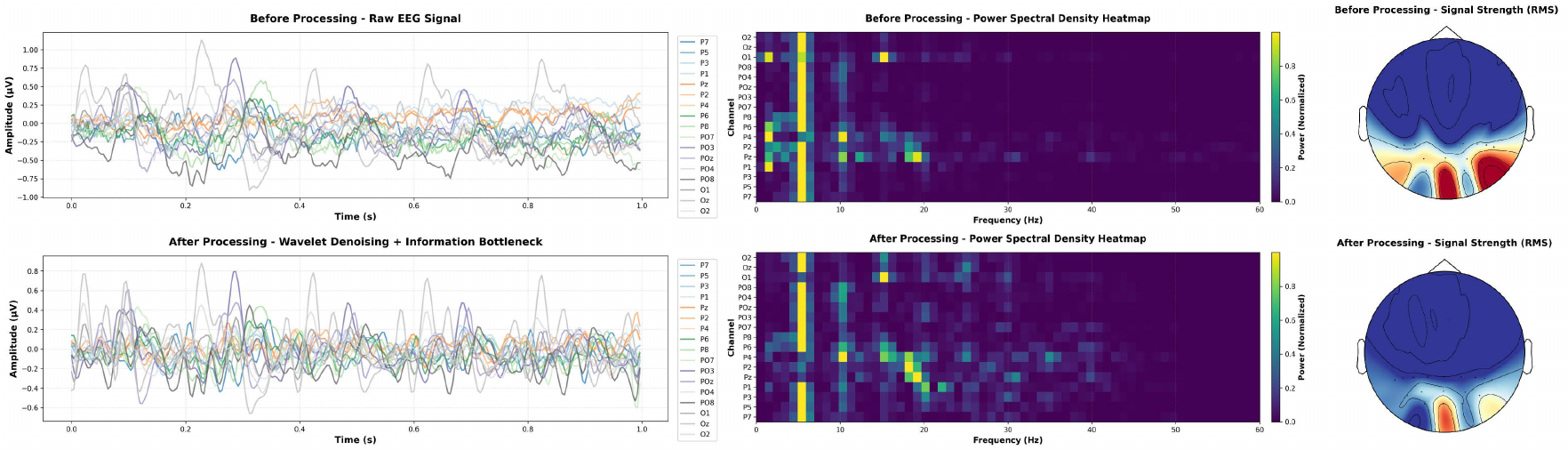}
  \caption{EEG signals before and after frequency screening in the subject-independent setting, shown in the time domain (left), frequency domain (middle), and EEG topographic map (right).}
  \label{inter_denoise}
\end{figure}

Under the subject-independent setting, spectral selection places greater emphasis on the Beta band (\cref{inter_scale}). The time-domain amplitude range changes from approximately $[-1, 1]$ to $[-0.6, 0.8]$, while components above 50 Hz are attenuated and activity in the 20--40 Hz range becomes more prominent (\cref{inter_denoise}). The posterior topographic pattern extends toward the right-parietal region, which may reflect contributions from attentional modulation and visual information integration. Compared with the subject-dependent visualization, the attenuation is less pronounced, suggesting that screening across subjects must accommodate more heterogeneous neural responses.

\subsection{Boundary Calibration Analysis}
\label{app:boundary_calibration}

\subsubsection{Distributional Diagnostics and Reference Boundaries}

We examine matched-pair similarity distributions separately for all ten subjects using Shapiro--Wilk tests and normal Q--Q plots (\cref{tab:distribution_diagnostics}). Normality is not rejected for subjects 03, 06, and 09 at the 5\% significance level, whereas the remaining subjects exhibit departures from a strict Gaussian distribution. Nevertheless, the squared Q--Q correlations exceed 0.96 for all subjects, supporting the Gaussian model as a reference approximation for boundary construction.

\begin{table}[htbp]
  \centering
  \caption{Per-subject distributional diagnostics on THINGS-EEG.}
  \label{tab:distribution_diagnostics}
  \resizebox{\linewidth}{!}{%
  \begin{tabular}{lcccccccccc}
    \toprule
    Metric & Sub 1 & Sub 2 & Sub 3 & Sub 4 & Sub 5 & Sub 6 & Sub 7 & Sub 8 & Sub 9 & Sub 10 \\
    \midrule
    Shapiro--Wilk $p$ & 0.0001 & 0.0007 & 0.7582 & 0.0054 & 0.0229 & 0.2535 & 0.0120 & 0.0010 & 0.1762 & 0.0003 \\
    Q--Q $R^2$ & 0.964 & 0.977 & 0.996 & 0.983 & 0.985 & 0.992 & 0.983 & 0.978 & 0.993 & 0.972 \\
    \bottomrule
  \end{tabular}%
  }
\end{table}

\begin{figure}[htbp]
  \centering
  \begin{subfigure}[t]{0.48\textwidth}
    \centering
    \includegraphics[width=\linewidth]{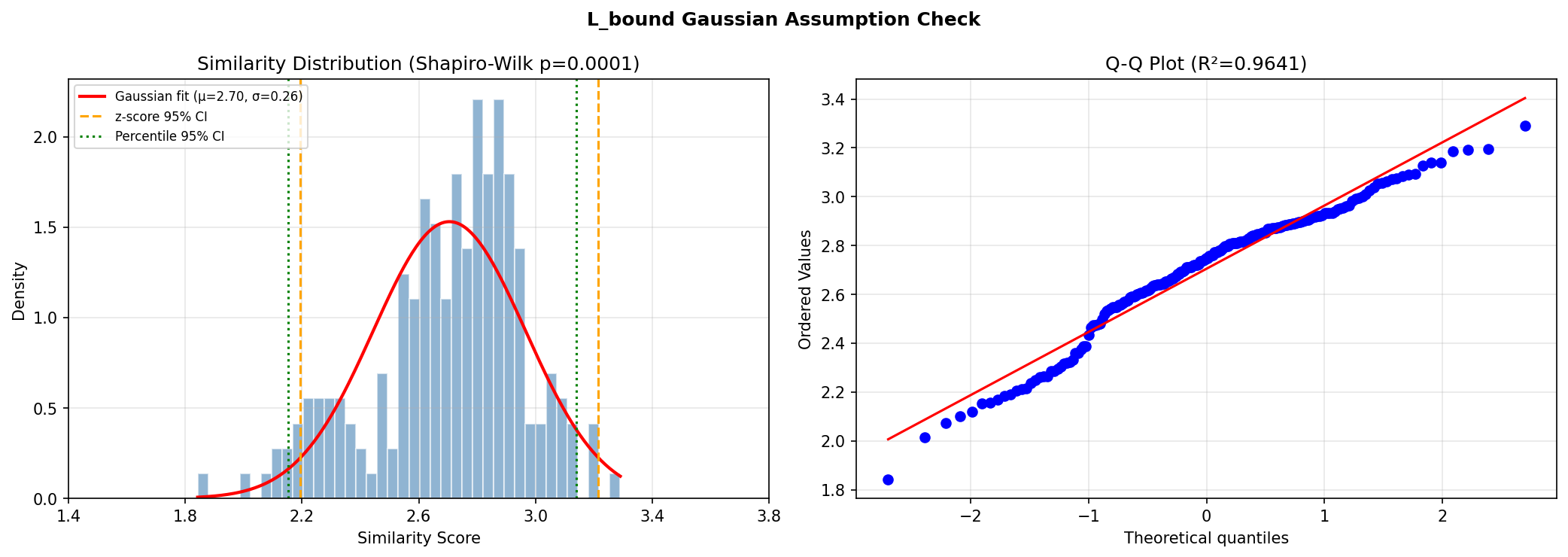}
    \caption{Subject 01.}
  \end{subfigure}
  \hfill
  \begin{subfigure}[t]{0.48\textwidth}
    \centering
    \includegraphics[width=\linewidth]{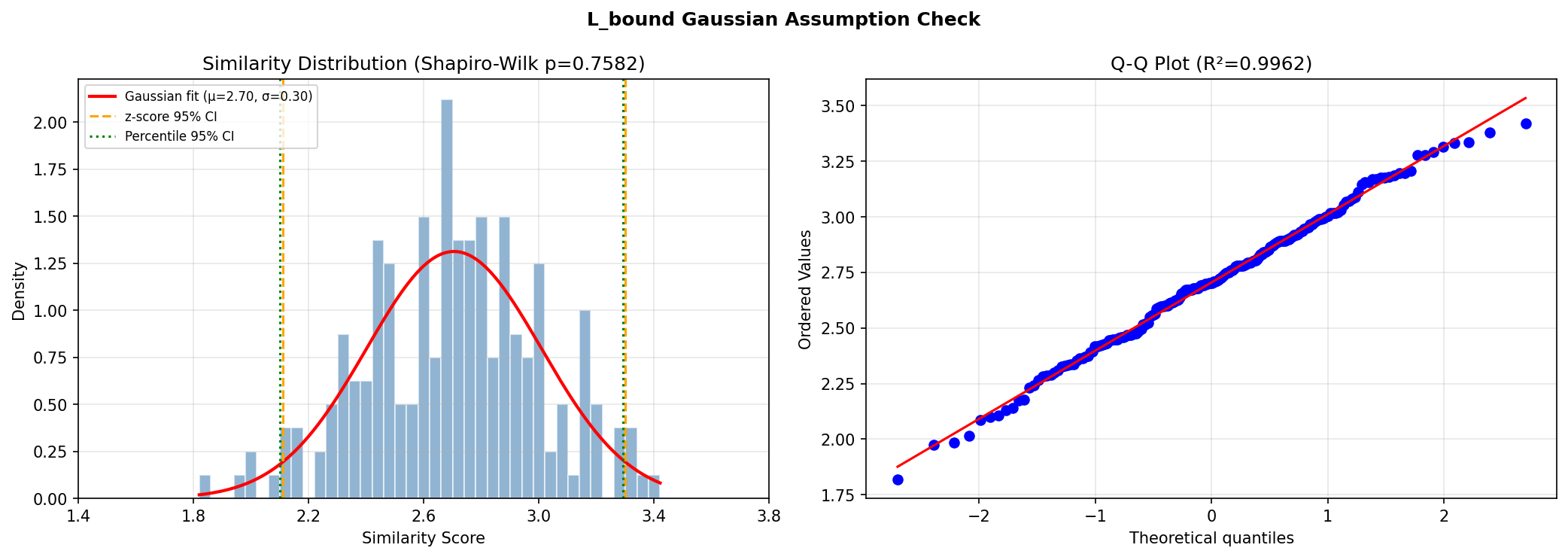}
    \caption{Subject 03.}
  \end{subfigure}

  \medskip

  \begin{subfigure}[t]{0.48\textwidth}
    \centering
    \includegraphics[width=\linewidth]{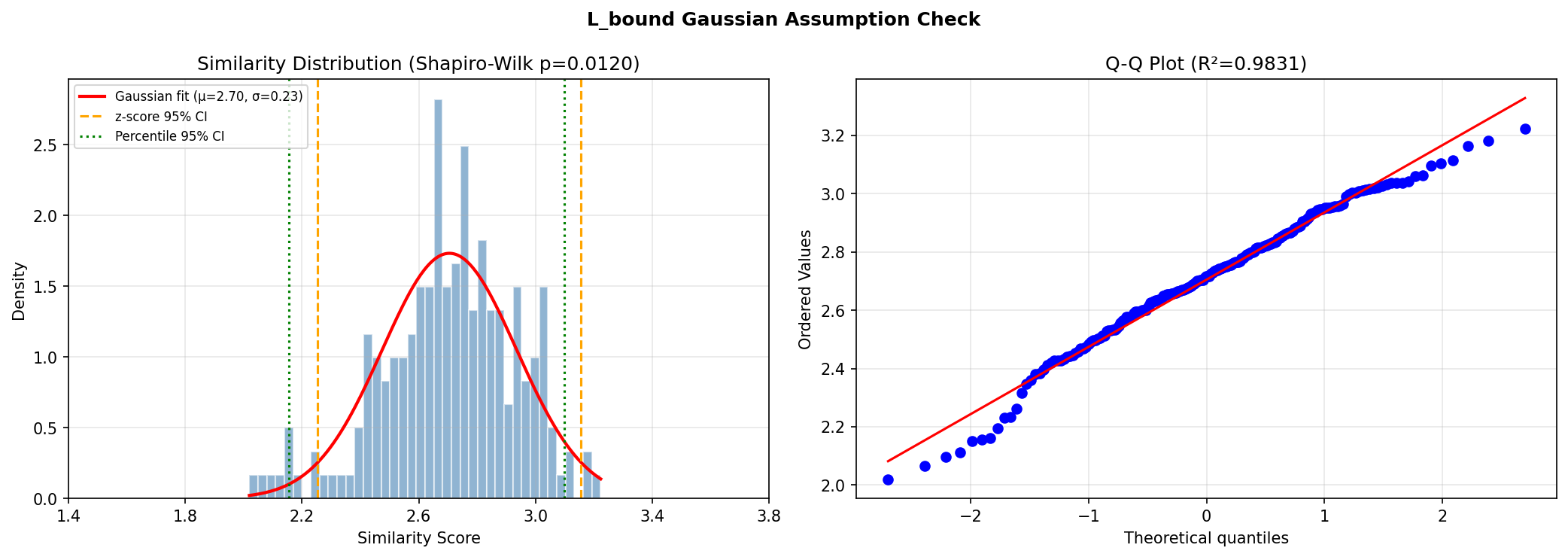}
    \caption{Subject 07.}
  \end{subfigure}
  \hfill
  \begin{subfigure}[t]{0.48\textwidth}
    \centering
    \includegraphics[width=\linewidth]{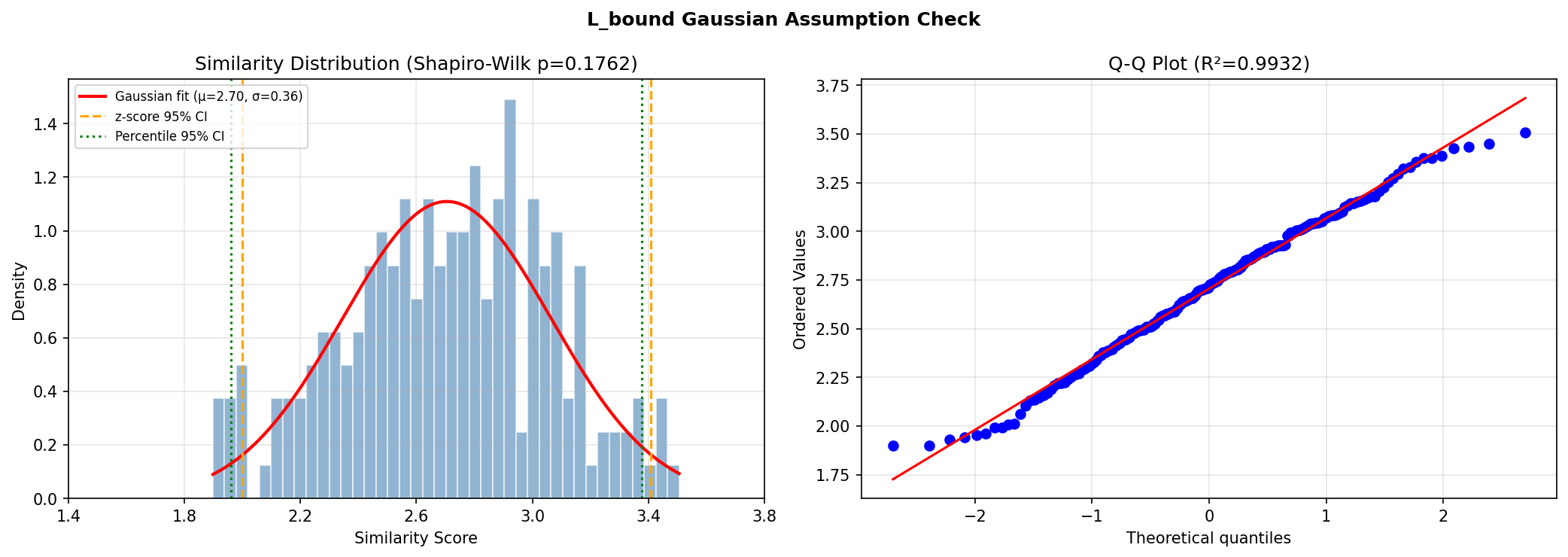}
    \caption{Subject 09.}
  \end{subfigure}
  \caption{Per-subject similarity distributions and normal Q--Q plots. Orange dashed and green dotted lines indicate Gaussian-reference and empirical percentile boundaries, respectively.}
  \label{fig:subject_gaussian_diagnostics}
\end{figure}

\cref{fig:subject_gaussian_diagnostics} shows representative distributions for subjects 01, 03, 07, and 09. Subjects 01 and 07 exhibit more pronounced deviations from the normal reference line, while subjects 03 and 09 follow it more closely. The Gaussian-reference and empirical percentile boundaries remain visually close in these examples, suggesting that moderate distributional deviations need not produce markedly different reference intervals.

\subsubsection{Comparison with Alternative Losses}

We compare boundary calibration with SupCon and hard-example weighting. These alternatives achieve 66.7\% / 94.6\% and 64.1\% / 93.0\% Top-1/Top-5 accuracy, respectively, below CAIA's 70.3\% / 96.1\% (\cref{tab:boundary_alternatives}). Together with the component ablations, these results support boundary calibration as a complementary signal for adaptive visual fusion and spectral screening.

\begin{table}[htbp]
  \centering
  \caption{Top-1/Top-5 accuracy (\%) for alternative losses on THINGS-EEG.}
  \label{tab:boundary_alternatives}
  \begin{tabular}{lcc}
    \toprule
    Method & Top-1 & Top-5 \\
    \midrule
    SupCon & \underline{66.7} & \underline{94.6} \\
    Hard-example weighting & 64.1 & 93.0 \\
    \rowcolor{blue!10}
    CAIA & \textbf{70.3} & \textbf{96.1} \\
    \bottomrule
  \end{tabular}
\end{table}

\subsubsection{Reference Interval Sensitivity}

We evaluate reference interval coverage levels of 90\%, 95\%, and 99\% (\cref{tab:alpha_sensitivity}). Mean Top-1 accuracy remains between 69.9\% and 70.3\%, while Top-5 accuracy ranges from 94.7\% to 96.1\%. The 95\% setting achieves the highest mean accuracy for both metrics, indicating limited sensitivity within the tested range.

\begin{table}[htbp]
  \centering
  \caption{Sensitivity to reference interval coverage on THINGS-EEG (200-way zero-shot retrieval, subject-dependent, Top-1/Top-5, \%).}
  \label{tab:alpha_sensitivity}
  \resizebox{\linewidth}{!}{%
  \begin{tabular}{lcccccccccccccccccccccc}
    \toprule
    & \multicolumn{2}{c}{Sub 1} & \multicolumn{2}{c}{Sub 2} & \multicolumn{2}{c}{Sub 3} & \multicolumn{2}{c}{Sub 4} & \multicolumn{2}{c}{Sub 5} & \multicolumn{2}{c}{Sub 6} & \multicolumn{2}{c}{Sub 7} & \multicolumn{2}{c}{Sub 8} & \multicolumn{2}{c}{Sub 9} & \multicolumn{2}{c}{Sub 10} & \multicolumn{2}{c}{Avg} \\
    \cmidrule(r){2-3} \cmidrule(r){4-5} \cmidrule(r){6-7} \cmidrule(r){8-9} \cmidrule(r){10-11} \cmidrule(r){12-13} \cmidrule(r){14-15} \cmidrule(r){16-17} \cmidrule(r){18-19} \cmidrule(r){20-21} \cmidrule(r){22-23}
    $1-\alpha$ & T1 & T5 & T1 & T5 & T1 & T5 & T1 & T5 & T1 & T5 & T1 & T5 & T1 & T5 & T1 & T5 & T1 & T5 & T1 & T5 & T1 & T5 \\
    \midrule
    99\% & 63.0 & 97.0 & \textbf{80.0} & \textbf{97.5} & \textbf{79.0} & \textbf{99.0} & 50.5 & 88.5 & \underline{63.0} & \underline{93.0} & \underline{77.5} & 95.0 & \textbf{64.5} & \underline{93.5} & \underline{80.0} & 97.0 & \underline{64.5} & \underline{94.0} & \underline{80.0} & \textbf{97.5} & \underline{70.2} & \underline{95.2} \\
    90\% & \underline{69.5} & \underline{97.5} & 78.5 & 96.5 & 69.0 & \underline{97.5} & \textbf{56.0} & \textbf{92.5} & 58.5 & 90.5 & \textbf{79.0} & \underline{95.5} & \underline{64.0} & 93.0 & 80.0 & \underline{97.5} & 63.5 & 90.0 & \textbf{80.5} & 96.0 & 69.9 & 94.7 \\
    \rowcolor{blue!10}
    95\% & \textbf{74.0} & \textbf{98.0} & \textbf{80.0} & \textbf{97.5} & \underline{69.0} & 94.0 & \underline{51.0} & \textbf{92.5} & \textbf{67.5} & \textbf{94.0} & 67.5 & \textbf{96.5} & 63.0 & \textbf{95.5} & \textbf{83.0} & \textbf{99.0} & \textbf{69.0} & \textbf{96.5} & 79.0 & \textbf{97.5} & \textbf{70.3} & \textbf{96.1} \\
    \bottomrule
  \end{tabular}%
  }
\end{table}

\subsection{Computational Cost Analysis}
\label{app:computational_cost}

We compare the computational cost of CAIA and UBP~\cite{t20} in \cref{tab:cost}. CAIA introduces additional parameters and FLOPs through adaptive visual fusion and frequency screening, with an inference latency of 14.29\,ms compared with 5.88\,ms for UBP.

Prior work discusses latency targets below 200\,ms for EEG decoding~\cite{mindcross} and tighter requirements in some implantable applications~\cite{infinimind}, while wireless EEG acquisition can itself introduce delays of tens of milliseconds~\cite{wireless_eeg}. These considerations place the additional model execution time in the context of a broader acquisition and decoding pipeline. CAIA's measured inference latency remains below these reported latency targets, suggesting potential for latency-sensitive decoding applications, although end-to-end feasibility depends on acquisition, preprocessing, and trial aggregation.

The additional training and inference costs accompany substantial retrieval gains, providing a practical trade-off between computational efficiency and decoding performance.

\begin{table}[htbp]
  \centering
  \caption{Computational cost comparison on an NVIDIA RTX~3090.}
  \label{tab:cost}
  \begin{tabular}{lcccc}
    \toprule
    Method & Params (M) & FLOPs (G) & Inference (ms) & Train Step (ms) \\
    \midrule
    UBP & 43.72 & 5.423 & 5.88 $\pm$ 0.58 & 8.35 $\pm$ 0.49 \\
    CAIA & 107.69 & 6.559 & 14.29 $\pm$ 0.44 & 34.79 $\pm$ 0.42 \\
    \bottomrule
  \end{tabular}
\end{table}

\subsection{Retrieval Examples and Embedding Visualizations}
\label{app:retrieval_visualizations}

\subsubsection{Retrieval Examples}

\begin{figure}[htbp]
  \centering
  \includegraphics[width=0.8\linewidth]{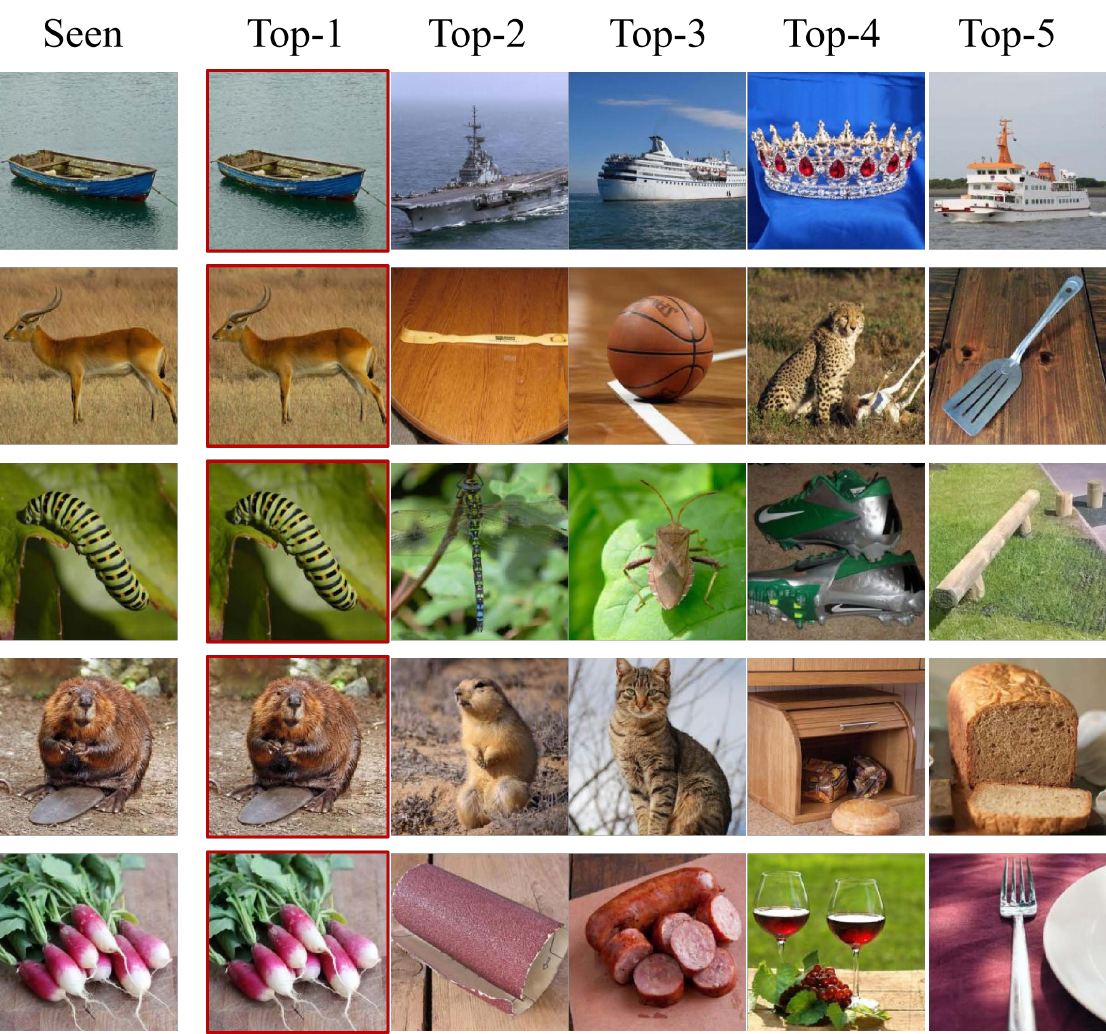}
  \caption{Successful retrieval examples.}
  \label{fig:retrieval_success}
\end{figure}

\cref{fig:retrieval_success} shows representative examples with correct Top-1 predictions, illustrating successful matching between EEG responses and their corresponding visual stimuli.

\begin{figure}[htbp]
  \centering
  \includegraphics[width=0.8\linewidth]{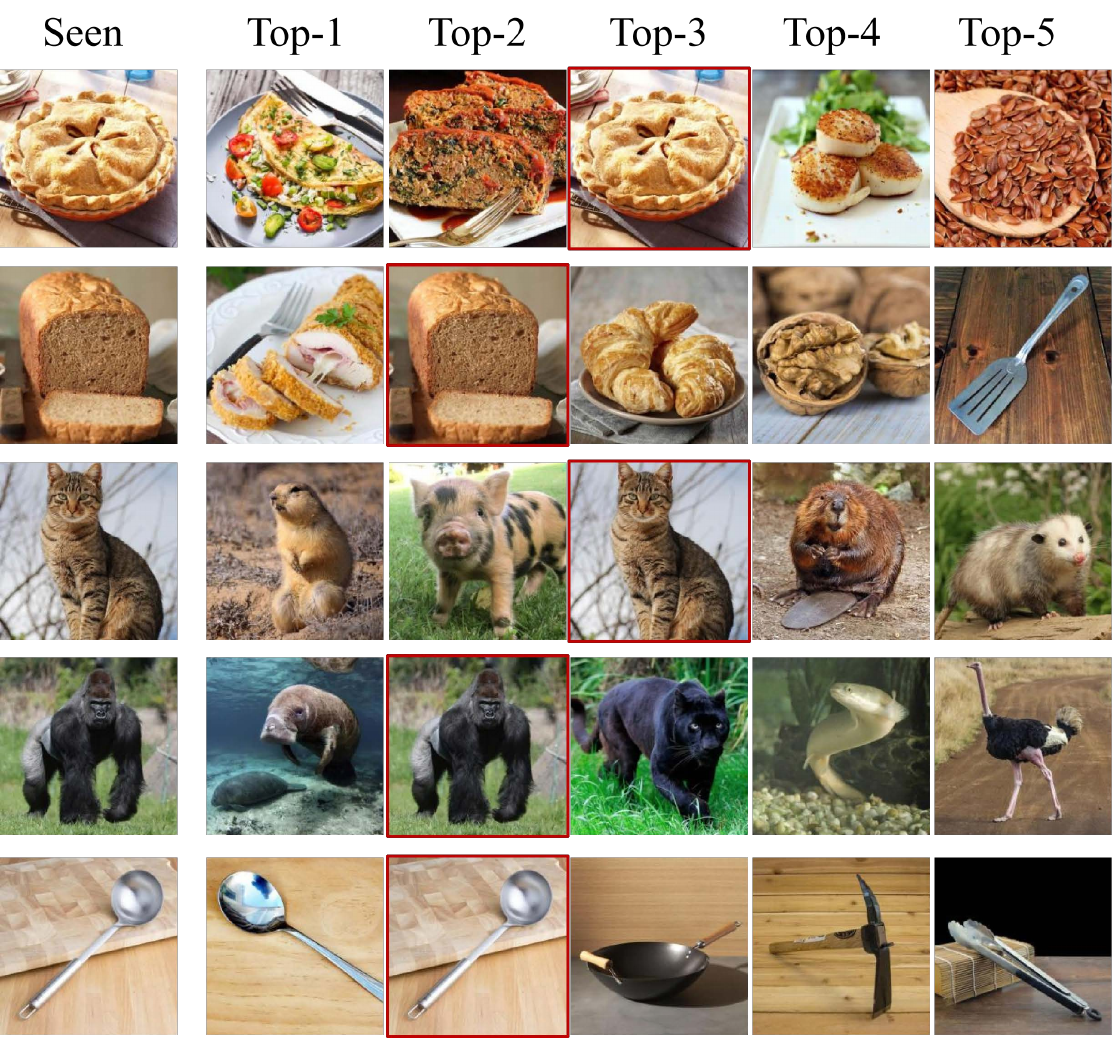}
  \caption{Failed retrieval examples.}
  \label{fig:retrieval_failure}
\end{figure}

\cref{fig:retrieval_failure} presents examples with incorrect Top-1 predictions. Visually similar or semantically related alternatives appear among the Top-5 candidates, suggesting that the learned representations retain useful category-level information even when the target is not ranked first. These cases highlight the remaining difficulty of distinguishing visually or semantically similar stimuli.

\subsubsection{UMAP Visualization and Stability}

\begin{figure}[htbp]
  \centering
  \begin{subfigure}[t]{0.32\textwidth}
    \centering
    \includegraphics[width=\linewidth]{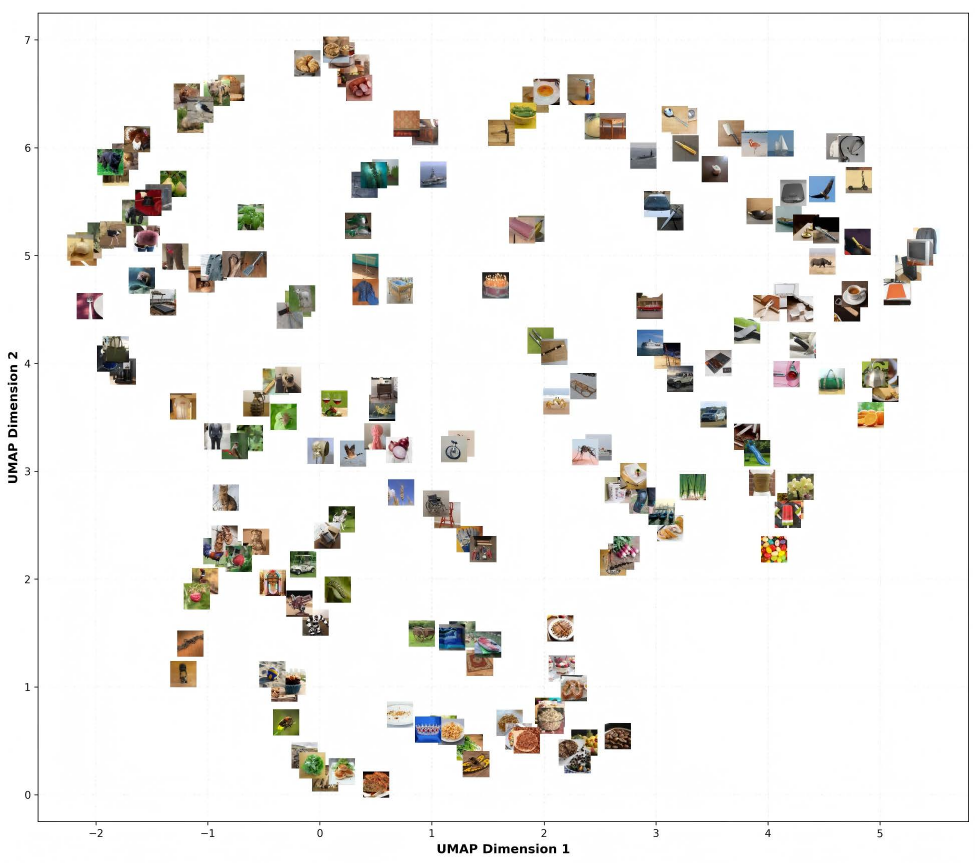}
    \caption{EEG representations.}
    \label{fig:umap_eeg}
  \end{subfigure}
  \hfill
  \begin{subfigure}[t]{0.32\textwidth}
    \centering
    \includegraphics[width=\linewidth]{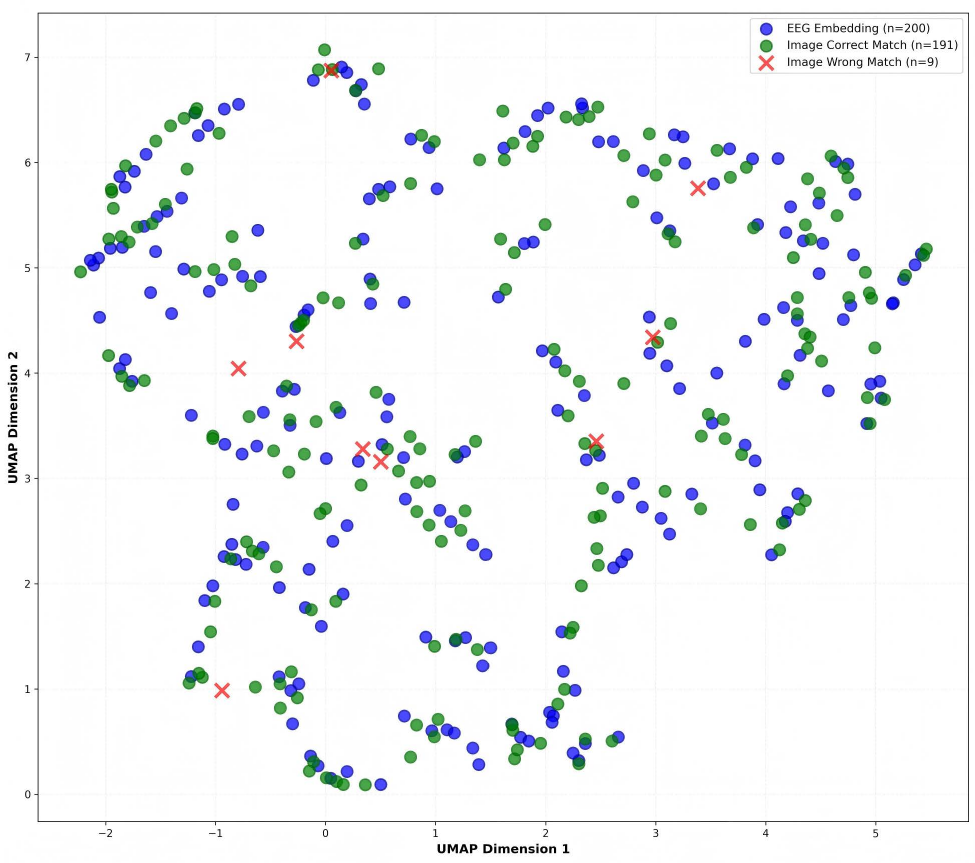}
    \caption{Joint representations.}
    \label{fig:umap_joint}
  \end{subfigure}
  \hfill
  \begin{subfigure}[t]{0.32\textwidth}
    \centering
    \includegraphics[width=\linewidth]{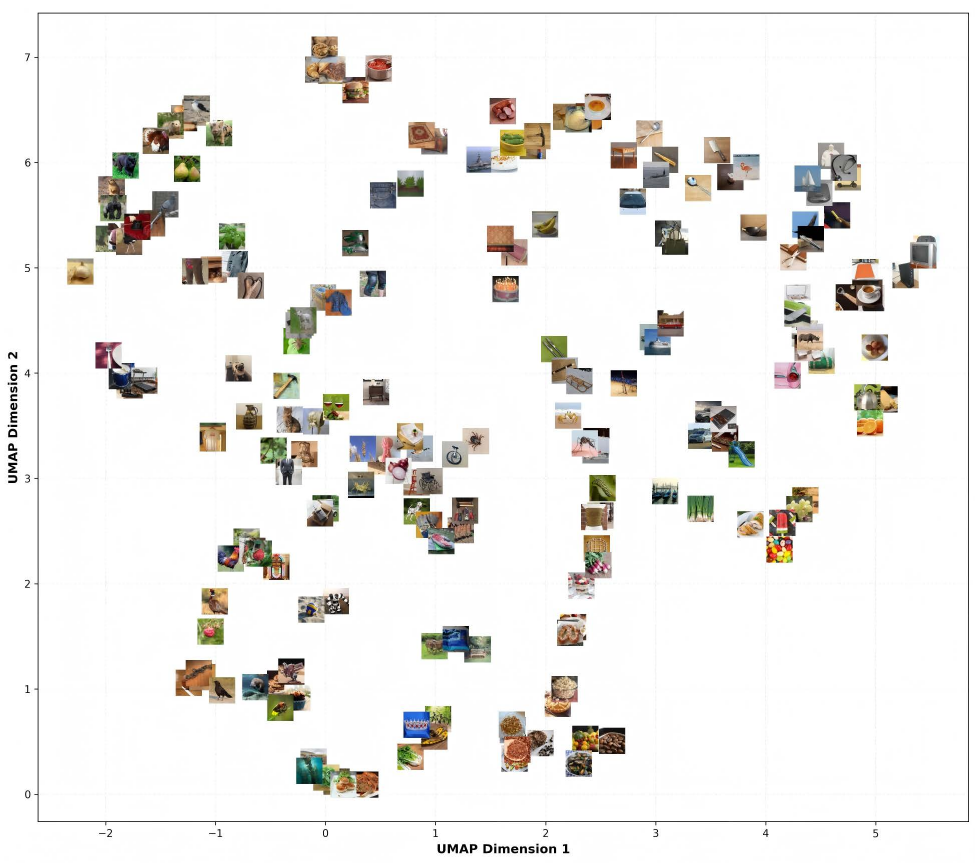}
    \caption{Image representations.}
    \label{fig:umap_image}
  \end{subfigure}
  \caption{UMAP visualizations of EEG and image representations. In (b), green and red indicate correct and incorrect retrievals, respectively.}
  \label{fig:three_subfigures}
\end{figure}

\cref{fig:three_subfigures} visualizes the EEG and image representations using UMAP. The modality-specific projections display broadly similar spatial organization, while the joint projection illustrates the distribution of correctly and incorrectly retrieved samples. These visualizations provide a qualitative view of the learned cross-modal representations.

To examine the stability of this geometric correspondence, we repeat the Procrustes analysis under five UMAP random seeds with $n_{\mathrm{neighbors}}=15$ and $\mathrm{min\_dist}=0.1$. The matched-pair disparity ranges from 0.108 to 0.121, with a mean of 0.115, compared with approximately 0.993 for the random baseline (\cref{tab:umap_stability}). These results suggest that the observed geometric correspondence is stable across the tested UMAP initializations.

\begin{table}[htbp]
  \centering
  \caption{Procrustes disparity across five UMAP random seeds.}
  \label{tab:umap_stability}
  \begin{tabular}{lccccc}
    \toprule
    UMAP Seed & 0 & 1 & 2 & 3 & 4 \\
    \midrule
    Matched pairs & 0.114 & 0.119 & 0.108 & 0.121 & 0.111 \\
    Random baseline & 0.993 & 0.993 & 0.994 & 0.992 & 0.993 \\
    \bottomrule
  \end{tabular}
\end{table}

\clearpage
\section*{NeurIPS Paper Checklist}

\begin{enumerate}

\item {\bf Claims}
    \item[] Question: Do the main claims made in the abstract and introduction accurately reflect the paper's contributions and scope?
    \item[] Answer: \answerYes{} 
    \item[] Justification: The abstract and introduction state the proposed CAIA framework and its components, supported by the experiments in \cref{sec:experiments} and the supplementary analyses in Appendices~\ref{app:robustness_capacity}--\ref{app:retrieval_visualizations}.

\item {\bf Limitations}
    \item[] Question: Does the paper discuss the limitations of the work performed by the authors?
    \item[] Answer: \answerYes{} 
    \item[] Justification: \cref{sec:conclusion} discusses limitations related to sustained central fixation, repeated-trial averaging, and the absence of eye-tracking-based validation.

\item {\bf Theory assumptions and proofs}
    \item[] Question: For each theoretical result, does the paper provide the full set of assumptions and a complete (and correct) proof?
    \item[] Answer: \answerNA{} 
    \item[] Justification: The paper does not include theoretical results; it focuses on empirical evaluation of the proposed method.

    \item {\bf Experimental result reproducibility}
    \item[] Question: Does the paper fully disclose all the information needed to reproduce the main experimental results of the paper to the extent that it affects the main claims and/or conclusions of the paper (regardless of whether the code and data are provided or not)?
    \item[] Answer: \answerYes{} 
    \item[] Justification: \cref{sec:method,sec:implementation} describe the model, datasets, preprocessing, and training settings. Additional dataset details are provided in Appendix~\ref{app:datasets}, and code is available through the repository linked in the abstract.

\item {\bf Open access to data and code}
    \item[] Question: Does the paper provide open access to the data and code, with sufficient instructions to faithfully reproduce the main experimental results, as described in supplemental material?
    \item[] Answer: \answerYes{} 
    \item[] Justification: Code is available at \url{https://github.com/mikumafu/CAIA}. THINGS-EEG and THINGS-MEG are publicly available and cited in \cref{sec:implementation} and Appendix~\ref{app:datasets}.

\item {\bf Experimental setting/details}
    \item[] Question: Does the paper specify all the training and test details (e.g., data splits, hyperparameters, how they were chosen, type of optimizer) necessary to understand the results?
    \item[] Answer: \answerYes{} 
    \item[] Justification: \cref{sec:implementation} details the data splits, preprocessing, optimizer, batch sizes, and learning rates for both subject-dependent and subject-independent settings.

\item {\bf Experiment statistical significance}
    \item[] Question: Does the paper report error bars suitably and correctly defined or other appropriate information about the statistical significance of the experiments?
    \item[] Answer: \answerYes{} 
    \item[] Justification: Appendix~\ref{app:robustness_capacity} reports standard errors and 95\% confidence intervals across five seed-level mean accuracies, together with two-sided paired $t$-tests across ten subjects for seed-matched CAIA and UBP runs. Confidence intervals use Student's $t$ distribution with four degrees of freedom.

\item {\bf Experiments compute resources}
    \item[] Question: For each experiment, does the paper provide sufficient information on the computer resources (type of compute workers, memory, time of execution) needed to reproduce the experiments?
    \item[] Answer: \answerYes{} 
    \item[] Justification: \cref{sec:implementation} identifies the NVIDIA RTX~3090 GPU. Appendix~\ref{app:computational_cost} and \cref{tab:cost} report model size, FLOPs, inference latency, and training step time for UBP and CAIA.
    
\item {\bf Code of ethics}
    \item[] Question: Does the research conducted in the paper conform, in every respect, with the NeurIPS Code of Ethics \url{https://neurips.cc/public/EthicsGuidelines}?
    \item[] Answer: \answerYes{} 
    \item[] Justification: The research uses publicly available datasets and adheres to standard ethical practices. No human subjects were directly involved. The paper does not violate any ethical guidelines.

\item {\bf Broader impacts}
    \item[] Question: Does the paper discuss both potential positive societal impacts and negative societal impacts of the work performed?
    \item[] Answer: \answerNo{} 
    \item[] Justification: This work is foundational research aimed at improving EEG‑visual alignment. It focuses on algorithmic design and empirical validation on public datasets, and does not directly address deployment‑level societal impacts.
    
\item {\bf Safeguards}
    \item[] Question: Does the paper describe safeguards that have been put in place for responsible release of data or models that have a high risk for misuse (e.g., pre-trained language models, image generators, or scraped datasets)?
    \item[] Answer: \answerNA{} 
    \item[] Justification: The paper does not release any high-risk models or datasets; it only introduces an alignment framework that does not inherently pose misuse risks.

\item {\bf Licenses for existing assets}
    \item[] Question: Are the creators or original owners of assets (e.g., code, data, models), used in the paper, properly credited and are the license and terms of use explicitly mentioned and properly respected?
    \item[] Answer: \answerYes{} 
    \item[] Justification: All used datasets and prior methods are cited. The datasets are publicly available research datasets; specific licenses are stated in their respective publications.

\item {\bf New assets}
    \item[] Question: Are new assets introduced in the paper well documented and is the documentation provided alongside the assets?
    \item[] Answer: \answerYes{} 
    \item[] Justification: The code repository linked in the abstract provides implementation documentation.

\item {\bf Crowdsourcing and research with human subjects}
    \item[] Question: For crowdsourcing experiments and research with human subjects, does the paper include the full text of instructions given to participants and screenshots, if applicable, as well as details about compensation (if any)? 
    \item[] Answer: \answerNA{} 
    \item[] Justification: The paper uses existing public datasets; no new data collection from human subjects was conducted.

\item {\bf Institutional review board (IRB) approvals or equivalent for research with human subjects}
    \item[] Question: Does the paper describe potential risks incurred by study participants, whether such risks were disclosed to the subjects, and whether Institutional Review Board (IRB) approvals (or an equivalent approval/review based on the requirements of your country or institution) were obtained?
    \item[] Answer: \answerNA{} 
    \item[] Justification: The paper does not involve primary human subjects research; it solely uses previously published datasets.

\item {\bf Declaration of LLM usage}
    \item[] Question: Does the paper describe the usage of LLMs if it is an important, original, or non-standard component of the core methods in this research? Note that if the LLM is used only for writing, editing, or formatting purposes and does \emph{not} impact the core methodology, scientific rigor, or originality of the research, declaration is not required.
    \item[] Answer: \answerNA{} 
    \item[] Justification: No LLMs were used as part of the core method; the research does not rely on LLMs in any significant way.

\end{enumerate}

\end{document}